\documentclass{article}

\usepackage[final,nonatbib]{neurips_2022}
\usepackage[numbers]{natbib}

\usepackage[utf8]{inputenc} 
\usepackage[T1]{fontenc}    
\usepackage{hyperref}       
\usepackage{url}            
\usepackage{booktabs}       
\usepackage{amsfonts}       
\usepackage{nicefrac}       
\usepackage{microtype}      
\usepackage{xcolor}         
\usepackage{multirow}

\usepackage{mathtools}
\usepackage{caption}
\usepackage{subcaption}
\usepackage{amsfonts}
\usepackage{mathrsfs}

\usepackage{amsmath}
\usepackage{amssymb}
\usepackage{centernot}

\author{%
  Dmitry Ivanov\thanks{\texttt{dimonenka@mail.ru, divanov@campus.technion.ac.il, diivanov@hse.ru}} \\
  HSE University \& Technion\\
  Israel \\
   \And
   Iskander Safiulin \\
   Independent researcher \\
   Russia \\
   \And
   Igor Filippov \\
   Myna Labs, Inc. \\
   USA \\
   \And
   Ksenia Balabaeva \\
   ITMO University \\
   Russia \\
}

\title{Optimal-er Auctions through Attention}

\begin{document}

\maketitle

\begin{abstract}
  RegretNet is a recent breakthrough in the automated design of revenue-maximizing auctions. It combines the flexibility of deep learning with the regret-based approach to relax the Incentive Compatibility (IC) constraint (that participants prefer to bid truthfully) in order to approximate optimal auctions. We propose two independent improvements of RegretNet. The first is a neural architecture denoted as RegretFormer that is based on attention layers. The second is a loss function that requires explicit specification of an acceptable IC violation denoted as regret budget. We investigate both modifications in an extensive experimental study that includes settings with constant and inconstant number of items and participants, as well as novel validation procedures tailored to regret-based approaches. We find that RegretFormer consistently outperforms RegretNet in revenue (i.e. is \textit{optimal-er}) and that our loss function both simplifies hyperparameter tuning and allows to unambiguously control the revenue-regret trade-off by selecting the regret budget.\footnote{Code is available \href{https://github.com/dimonenka/optimaler}{here}}
\end{abstract}

\section{Introduction}

Several decades ago, \citet{myerson1981optimal} has proposed a general solution for the problem of optimal (revenue-maximizing) design of single-item auctions. Despite the collaborative effort of the community, the results of the same generality for the multi-item auctions have not been obtained. As an alternative to analytical solutions, automated auction design \cite{conitzer2002complexity,conitzer2004self} takes a perspective of constrained optimization. In particular, the objective is to maximize the revenue while satisfying the Incentive Compatibility (IC) and Individual Rationality (IR) constraints. This framework can provide approximate solutions in settings where the optimal mechanisms are unknown, as well as hint at what the analytical solutions may look like. The early approaches had utilized linear programming and classic machine learning. However, after the renaissance of deep learning, combining auction design with neural networks was only a question of time.

\citet{dutting2019optimal} are the pioneers of this approach with their RegretNet architecture. RegretNet is a neural network that represents a mechanism by mapping a matrix of bids that $n$ participants report for $m$ items into a probabilistic allocation matrix and a payment vector. It is trained via differential optimization on a mixture of two objectives: to maximize revenue and to minimize regret, which is a relaxation of the IC constraint proposed in \cite{dutting2015payment}. RegretNet recovers near-optimal revenue in the analytically solved settings and outperforms the previous state-of-the-art while having vanishing regret guarantees. In this paper, we propose two independent improvements of RegretNet.

\begin{figure}[t]
    \centering
    \includegraphics[width=\linewidth]{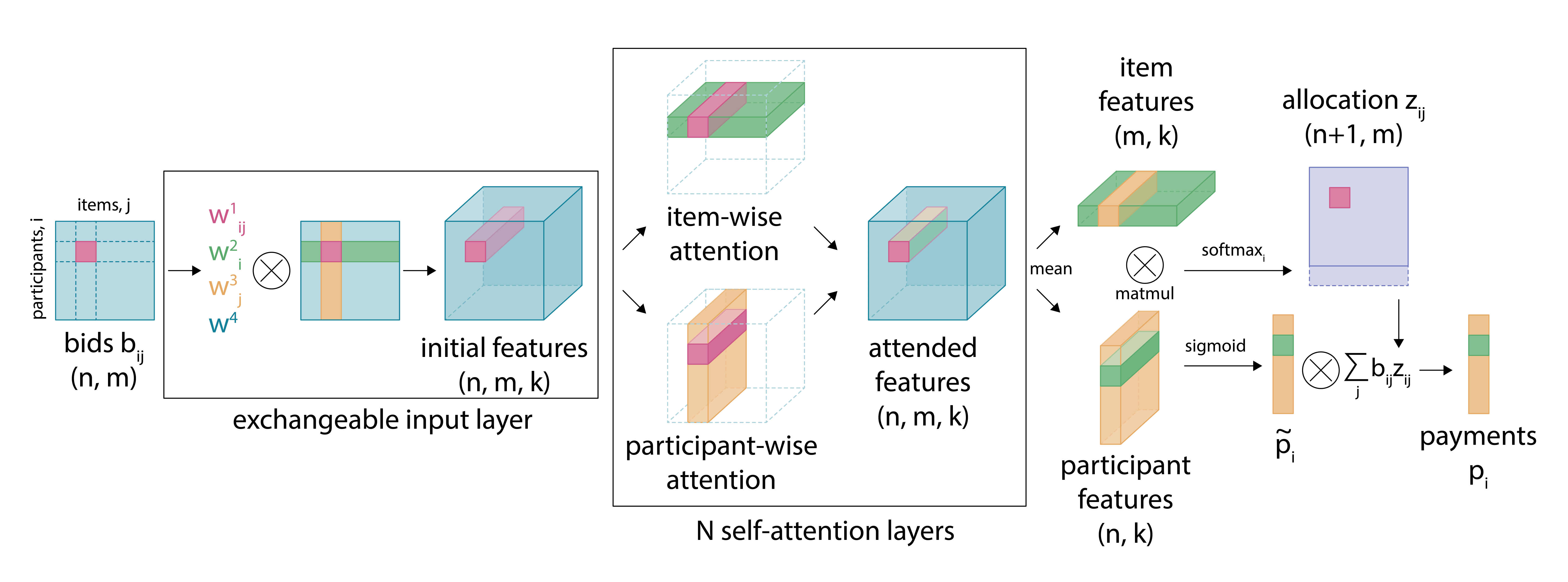}
    \caption{Our architecture RegretFormer. The description is provided in Section \ref{sec:our_architecture}.}
    \label{fig:scheme}
\end{figure}

First, we propose a neural architecture based on attention layers. We name it RegretFormer after the widely-known Transformers \cite{vaswani2017attention}. An illustration of the architecture is provided in Figure \ref{fig:scheme}. RegretFormer has several crucial properties. On the one hand, it is by design insensitive to the order of items and participants, which is desirable for learning symmetric auctions. Note that it can also be modified to learn asymmetric mechanisms. On the other hand, it does not assume a predetermined constant input size, which allows one network to learn from data consisting of auctions of varying sizes, as well as to generalize to settings unseen during training.

Second, we propose an alternative objective. Instead of optimizing arbitrary mixtures of revenue and regret, our approach maximizes revenue given an explicitly specified regret budget, which is an acceptable violation of the IC constraint. This is technically implemented through the method of Lagrange multipliers and dual gradient decent. Our approach allows to unambiguously balance revenue and regret by varying the regret budget. As a bonus, we find that our loss function is significantly less sensitive to the choice of loss-related hyperparameters.

We conduct an extensive experimental study to verify the strengths of our modifications. Specifically, we compare our and existing architectures in both symmetric and asymmetric settings with constant input size, as well as in novel settings with inconstant input size. We find that RegretFormer consistently outperforms existing architectures in revenue given the same regret budget. We additionally verify this result with two novel validation procedures. The first procedure is based on estimating the regret of one network given the optimal misreports approximated by another network. The second procedure is based on network distillation. Regarding our loss modification, we confirm that specifying the regret budget results in the intended revenue-regret trades-off.

\section{Background}

\subsection{Problem statement}

Let \(N=\{1,..., n\}\) be a set of \(n\) bidders, \(M=\{1,..., m\}\) be a set of \(m\) items. Each bidder \(i\) evaluates all items with valuation function \(v_i : 2^M \to \mathbb{R} \). In our study, we only consider \emph {additive} valuations, meaning that for each item \(j\in M\) a bidder has a valuation function \(v_i({j})\), and a valuation of a subset \(S\subseteq M\) equals to the sum of valuations of the items in the subset \(v_i(S)=\Sigma _{j\in S} v_i({j})\). The valuation function can also be extended to evaluate probabilistic assignments: \(v_i(z_{ij})=z_{ij} v_i(j)\), where \(z_{ij}\) is the probability that participant \(i\) is assigned item \(j\).

The valuation function $v_i$ is drawn independently from the distribution \(F_i\) over all possible valuation functions \(V_i\). A set of valuations of all bidders forms a valuation profile \(v = (v_1, ..., v_n)\). An auctioneer does not know the bidders' valuations nor their distribution \(F = (F_1,...,F_n)\), but has a sample \( L = ({v^1,...,v^{|L|}})\) of valuation profiles drawn from \(F\). The bidders independently report their bids \(b = (b_1, ...,b_n) \in V\), based on which the auction allocates the items and charges payments. Formally, the auction is defined as a tuple \((g=(g_1,...,g_n), p=(p_1,...,p_n))\) of probabilistic allocation rules \(g_i : V \to [0, 1]^M \) and payment rules \(p_i : V \to \mathbb{R}_\geq\).

Define the bidder's utility as \(u_i(v_i, b) = v_i(g_i(b)) - p_i(b)\). Define the valuation profile without the valuation \(v_i\) as \(v_{-i}\). Similarly, define the bids without \(b_i\) as \(b_{-i}\), and the valuation profile of all bidders except \(i\) as \(V_{-i} = \Pi_{k\ne i}V_k\). An auction is \emph {dominant strategy incentive compatible} (DSIC) if the utility of each bidder is maximized by reporting their true valuations regardless of bids of other bidders: \(\forall i, v, b: u_i(v_i, v) \geq u_i(v_i, (b_i, v_{-i}))\). An auction is \emph {ex-post individually rational} (IR) if each agent's utility is non-negative for all possible valuations and bids: \(\forall i, v, b: u_i(v_i, (b_i, v_{-i}))\geqslant 0\).

Define the revenue of a profile as a sum of bidders' payments \(\Sigma_i p_i(v)\). The goal of the optimal auction design is to maximize the expected revenue subject to DSIC and IR constraints. The problem is analytically solved for the auctions with one item in the seminal work of \citet{myerson1981optimal}, provided the valuation distributions are known. There are no general solutions for auctions with $m > 1$.

The task of optimal auction design can be viewed as an optimization or a machine learning problem. In this case, the expected revenue takes the place of the objective. There is a class of auctions \((g(w), p(w)) \in M\) parameterized with \(w \in \mathbb{R}^d\) for some \(d \in \mathbb{N}\) and a set of bidder valuation profiles \(L\) drawn as independent and identically distributed variables from \(F\). Then, the goal of the optimization procedure is to find an auction that minimizes the negated expected revenue \(-\mathbb{E}_F[\Sigma_{i\in N} p_i(v;w)]\) while satisfying DSIC and IR constraints.

\subsection{RegretNet}

RegretNet was proposed by \citet{dutting2019optimal} as a deep learning solution for the optimal auction design. It can be used for multi-bidder and multi-item settings, the analytical solutions for which are unknown. RegretNet explicitly encodes the allocation and payment rules of the auction mechanism into weights of a neural network. These rules are discovered as a result of optimization procedure.

The architecture of RegretNet consists of two networks: the allocation network and the payment network. Both networks take flattened constant-sized bid matrix \(B^{nm}\) as an input and process it through several fully-connected layers.

The allocation network maps a matrix of bids to categorical distributions of allocations between participants for each item: \(A_{net}(B^{nm})=Z^{nm}\), where \(z_{ij}\) is the probability of allocating the \(j\)-th item to the \(i\)-th bidder. To allow for the possibility of an item remaining unallocated, the output of the final layer has the size of $(n+1) \cdot m$, imitating an additional dummy participant with zero valuations. Then, to obtain properly scaled allocation probabilities over participants for each item, the softmax activation is applied to the output of the final layer corresponding to each item. Specifically, $z_{ij} = e^{\tilde{z}_{ij}} / \sum_{i = 1}^{n+1} e^{\tilde{z}_{ij}}$, where $\tilde{z}_{ij}$ denotes the unnormalized allocation probability (i.e. logit).

The payment network maps a matrix of bids to a vector of pay values: \(P_{net}(B^{nm})=\hat{P}^n\), where \(\hat{p}_{i}\) is the fraction of the \(i\)-th bidder expected utility that the bidder transfers to the auctioneer. The payment is then computed as \(p_i=\hat{p_i}\stackrel{~}\sum_{j=1}^mz_{ij}b_{ij}\) for \(i=1, \dots, n\). To properly scale $\hat{p}_{i} \in [0, 1]$, a sigmoid activation is applied to the output of the final layer. Because the payment cannot exceed the bidder's expected utility, the mechanisms discovered by RegretNet always satisfy the IR constraint.

Given a sample of profiles $L$, RegretNet aims to optimize the following empirical objective:

\begin{equation}
\label{eq:objective}
    \begin{split}
        \min_w \hspace{5pt} & -\frac{1}{\left|L\right|} \sum_{l \in L} \sum_{i \in N} p_i(v^l; w) \\
        s.t. \hspace{5pt} & rgt_i(v^l; w) = 0, \hspace{5pt} \forall i \in N, l \in L
    \end{split}
\end{equation}

\noindent where $rgt_i$ denotes the $i$-th bidder's ex-post regret. Regret is a quantifiable relaxation of the DSIC constraint first introduced in \cite{dutting2015payment}. It is defined as the difference between the $i$-th bidder's utilities given the optimal misreport (that provides the highest utility) and the truthful bid:

\begin{equation}
\label{eq:regret}
    rgt_i(v^l; w) = \max_{v_i'} \left( u(v_i^l, (v_i', v_{-i}^l); w) - u(v_i^l, v^l; w) \right)
\end{equation}

To solve the constrained optimization problem \ref{eq:objective} over the space of the network parameters $w$, the authors of RegretNet employ the augmented Lagrangian method. This yields a loss function that combines revenue maximization with a penalty for violations of DSIC constraints:

\begin{equation}
\label{eq:loss_outer}
    \mathcal{L}_{outer}(w) = \sum_{i \in N} \left[ -P_i + \lambda_i \widetilde{R}_i + \frac{\rho}{2} \widetilde{R}_i^2 \right]
\end{equation}

\noindent where $B$ denotes mini-batch, $P_i = \frac{1}{\left|B\right|} \sum_{l \in B} p_i(v^l; w)$ is the average revenue from the $i$-th participant, $\widetilde{R}_i = \frac{1}{\left|B\right|} \sum_{l \in B} \widetilde{rgt}_i(v_i^{\prime l}, v^l; w)$ is the average approximate regret of the $i$-th participant, and $\widetilde{rgt}_i(v_i^{\prime l}, v^l; w) = \left( u(v_i^l, (v_i', v_{-i}^l); w) - u(v_i^l, v^l; w) \right)$ is an approximation of $rgt_i$. To find the approximate regret, the optimal misreports $v_i^{\prime l}$ are estimated via gradient descent in the inner optimization loop by minimizing the $i$-th participant negated utility as a function of their misreport:

\begin{equation}
\label{eq:loss_inner}
    \mathcal{L}_{inner}(v_i^{\prime l}) = -u(v_i^l, (v_i^{\prime l}, v_{-i}^l); w)
\end{equation}

The Lagrange multipliers $\lambda_i$ and $\rho$ are periodically updated throughout the training according to the schedules: $\lambda_i \leftarrow \lambda_i + \rho \widetilde{R}_i$ and $\rho \leftarrow \rho + \rho_\Delta$. The starting values of $\lambda_i$ and $\rho$ and the learning rate $\rho_\Delta$ are hyperparameters that control the trade-off between revenue and regret of the learned mechanism.

\subsection{EquivariantNet}

EquivariantNet is an alternative architecture proposed by \citet{rahme2021permutation} to effectively deal with symmetric auctions. It relies on a theorem originally proven by \cite{daskalakis2012symmetries} that there exists an optimal solution for a symmetric auction that is permutation-equivariant, i.e. insensitive to the order of inputs.

Like RegretNet, EquivariantNet has an allocation network and a payment network. Both networks consist of compositions of exchangeable layers that preserve permutation-equivariance \cite{hartford2018deep}. The exchangeable layer adds to each element $i,j$ of a matrix an aggregate statistic (like sum) over the row $i$, the column $j$, and the whole matrix, weighted by learnable parameters that are shared among all matrix elements. The precise definition is provided in the Appendix. The authors observe that EquiavariantNet produces competitive results but does not outperform RegretNet in revenue.

\subsection{Related work}

Early work that explores the automated solutions to auction design formulates the problem as linear program \cite{conitzer2002complexity,sandholm2003automated,conitzer2004self} or searches within specific families of DSIC auctions \cite{likhodedov2005approximating,sandholm2015automated}. The former approach suffers from scalability issues due to the curse of dimensionality \cite{guo2010computationally}, while the latter approach searches within a limited family of auctions that might not contain the optimal mechanism. Classic machine learning has also been applied to the auction design \cite{lahaie2011kernel,dutting2015payment,narasimhan2016automated}, but these approaches are considered less flexible and general than the deep learning alternatives \cite{dutting2019optimal}.

In recent years, multiple extensions of RegretNet have been proposed, including the extensions to the optimal auction design with budget constraints \cite{feng2018deep}, fairness constraints \cite{kuo2020proportionnet}, and human preferences over desirable allocations \cite{peri2021preferencenet}, as well as the problem of faculty allocation \cite{golowich2018deep} and the matching problem \cite{ravindranath2021deep}. \citet{shen2019automated} and \citet{dutting2019optimal} propose alternatives to RegretNet that are exactly DSIC but that are only applicable to auctions with one bidder. \citet{curry2022differentiable} combine the approach of searching within limited families of DSIC auctions with deep learning. \citet{rahme2021auction} propose a simplified loss function for RegretNet that is easier to tune, as well as a potentially faster regret estimation procedure based on training an adversarial bidder networks. Deep learning has been applied to other aspects of mechanism design, such as iterative combinatorial auctions \cite{weissteiner2020deep}, minimizing economic burden on bidders within the Groves family of auctions \cite{tacchetti2019neural}, optimal bidding strategies \cite{nedelec2021adversarial,nedelec2019learning}, optimal redistribution mechanisms \cite{manisha2018learning}, and E-commerce advertising \cite{liu2021neural}. Several studies have combined reinforcement learning with auction design \cite{tang2017reinforcement,jin2018real,nguyen2020bandit,ai2022reinforcement}. A plethora of research has also focused on the questions of sample complexity of designing revenue-maximizing auctions \cite{cole2014sample,morgenstern2015pseudo,dhangwatnotai2015revenue,balcan2016sample,morgenstern2016learning,mohri2016learning,narasimhan2016general,syrgkanis2017sample,gonczarowski2017efficient,huang2018making,hartline2019sample,guo2019settling,gonczarowski2021sample}.

\section{Our modifications of RegretNet}\label{sec:our}

In this section, we describe our two modifications of RegretNet. The first modification is a novel neural architecture for the optimal auction design, which we denote as RegretFormer. The second modification is an alternative constrained objective that yields a more convenient loss function.

\subsection{RegretFormer: enhancing RegretNet with attention layers}\label{sec:our_architecture}

The neural architecture of RegretNet has several issues. One issue is the sensitivity of RegretNet to the order of items and participants in the input bid matrix. Such order should not affect the outcome of a symmetric auction \cite{daskalakis2012symmetries,qin2022benefits}. Even in non-symmetric auctions, permutation-equivariant solutions sometimes achieve near-optimal revenue \cite{hartline2009simple,alaei2019optimal,jin2019tight,jin2020tight,kothari2019approximation}. This makes the option to wire equivariance to input order into the architecture desirable. Another issue is that RegretNet assumes a constant number of participants and items. This severely limits its practical applicability due to inability to generalize to unseen input sizes and learn from data with heterogenous input size. Finally, the fully-connected layers used in RegretNet have limited expressivity and may not have the right inductive biases for auction design. Taking these issues into account, we propose \emph{RegretFormer}.

The architecture of RegretFormer is illustrated in Figure \ref{fig:scheme}. The network takes the matrix of bids \(B^{nm}\)  as the input. First, we apply an exchangeable layer \cite{hartford2018deep} to transform each bid into an initial vector of features that contains information about other bids. Then, we apply item-wise (i.e. row-wise) and participant-wise (i.e. column-wise) self-attention layers \cite{vaswani2017attention} to each feature vector corresponding to each bid. For a given bid, the outputs of the two self-attention layers are transformed into a single feature vector through a fully-connected layer. These self-attention transformations can be applied several times. Finally, by averaging over rows and columns we transform the output of self-attention layers into the allocation matrix $Z^{nm}$ and the payment vector $P^n$. Note that each layer shares the parameters across all bids to ensure that the network is insensitive to the order of bids and is applicable to different input sizes. The detailed definitions and order of all layers are provided in the Appendix.

Our architecture has several advantages. On the one hand, it leverages the expressivity of attention layers that help to achieve the state-of-the-art performance in many diverse and complex problems, e.g. \cite{vaswani2017attention,devlin2018bert,yuan2021vit}. We empirically demonstrate superiority of RegretFormer in Section \ref{sec:experiments}. On the other hand, it maintains the equivariance and the invariance to the order of items and participants. The former forces the resulting mechanisms to be symmetric, which drastically reduces the solution search space. The latter allows RegretFormer to learn from data with inconstant input sizes, as well as to generalize to unseen settings. Note that RegretFormer can still learn asymmetric mechanisms by utilizing positional encoding, which we demonstrate empirically.

We note that a concurrent work by \citet{duan2022context} also combines Transformers and RegretNet. However, their focus is on integrating contextual information of bidders and items into the framework, whereas we perform a wider and more accurate comparison of neural architectures.

\subsection{Specifying regret budget}\label{sec:our_loss}

Both RegretNet and EquivariantNet optimize a mixture of two conflicting objectives (\ref{eq:loss_outer}), namely revenue maximization and regret minimization, and control their trade-off with hyperparameters like initial values and schedules of the Lagrange multipliers. There are two issues with this approach. The first issue is sensitivity. These hyperparameters have to be precisely tuned for each experiment, and as \citet{rahme2021auction} show, may massively degrade the performance if selected improperly. The second issue is uninterpretability. While increasing any of $\lambda_i, \rho, \rho_\Delta$ tightens the regret budget of the learned mechanism, the exact effect of such changes on resulting revenue and regret is unpredictable. Furthermore, the recipe for hyperparameters selection in new settings is unclear. \citet{rahme2021auction} propose a mixture of objectives that mitigates the first issue, but the second issue remains unresolved.

We propose an alternative perspective that mitigates both issues. Instead of balancing the two objectives, we maximize the revenue given a maximal regret budget $R_{max}$, which is pre-specified by the designer. This corresponds to a relaxed version of the constrained objective (\ref{eq:objective}):

\begin{equation}
\label{eq:objective_ours}
    \begin{split}
        \min_w \hspace{5pt} -&\frac{1}{\left|L\right|} \sum_{l \in L} \sum_{i \in N} p_i(v^l; w) \\
        s.t. \hspace{5pt} & \frac{1}{\left|L\right|} \sum_{l \in L} \sum_{i \in N} rgt_i(v^l; w) \leq R_{max}
    \end{split}
\end{equation}

Instead of the constrained objective (\ref{eq:objective_ours}), we optimize its dual by introducing the Lagrange multiplier $\gamma$ (note that $R_{max}$ does not depend on $w$ and thus can be dropped):

\begin{equation}
\label{eq:loss_outer_ours}
    \mathcal{L}_{outer}(w) = -\sum_{i \in N} P_i + \gamma \sum_{i \in N} \widetilde{R}_i
\end{equation}

\noindent Critically, unlike the original RegretNet we do not hand-select the Lagrange multiplier. Instead, $\gamma$ is dynamically updated to enforce the constraint satisfaction while exhausting all available regret budget $R_{max}$. To this end, we employ dual gradient descent \cite{boyd2004convex}. Specifically, we iterate between one gradient update of the network parameters $w$ to minimize (\ref{eq:loss_outer_ours}) and one update of $\gamma$ according to:

\begin{equation}
\label{eq:lm_update}
    \gamma \leftarrow \max \left( 0, \gamma + \gamma_\Delta \left( \log( \sum_{i \in N} \widetilde{R}_i / \sum_{i \in N} P_i) - \log(R_{max}) \right) \right)
\end{equation}

\noindent where $\gamma_\Delta$ is the learning rate for the dual variable. For convenience, in (\ref{eq:lm_update}) we normalize the regret estimate by revenue. This way, $R_{max} \in [0, 1]$ specifies the regret budget as a percentage of revenue. We also apply logarithms to both terms of the difference in (\ref{eq:lm_update}) for faster convergence of $\gamma$.

Additionally, we apply a decreasing schedule to $R_{max}$. If the regret budget $R_{max}$ is too tight at the beginning of training, the network may fail to escape the local optima of low revenue and zero regret. To avoid that, we initialize $R_{max}$ at a higher value $R_{max}^{start}$ and exponentially anneal it to the desirable budget $R_{max}^{end}$ throughout the training. This leads the network to first find the solutions with high revenue and then tighten the regret. This results in the following update rule for the regret budget:

\begin{equation}
    R_{max} \leftarrow \max \left( R_{max}^{end}, R_{max}^{mult} \cdot R_{max} \right)
\end{equation}

\noindent where $R_{max}^{mult} < 1$ controls the speed of annealing of $R_{max}$ to $R_{max}^{end}$.

Another example of applying dual gradient descent to enforce a constraint can be found in \cite{peng2018variational}.

The proposed approach has several advantages. On the one hand, it resolves the dichotomy of two conflicting objectives. While the regret budget needs to be explicitly set based on the designer's preferences, all other hyperparameters are then straightforward to tune by maximizing the revenue given the specified regret budget. This is unlike the original objective of RegretNet (\ref{eq:loss_outer}) where the regret budget is chosen implicitly through specifying uninterpretable loss-related hyperparameters like $\lambda_i$, $\rho$, $\rho_\Delta$. On the other hand, our approach is also significantly less sensitive to the loss-related hyperparameters. We empirically demonstrate this by using the same hyperparameters related to the budget $R_{max}$ and its schedule in all our experiments, in contrast to the objective (\ref{eq:loss_outer}) that requires a uniquely tuned set of loss-related hyperparameters to perform well in a given setting \cite{rahme2021auction}.

\section{Experiments}\label{sec:experiments}

In this section, we empirically investigate our modifications of RegretNet. We compare our RegretFormer with RegretNet and EquivariantNet in settings with constant number of participants and items used by \citet{dutting2019optimal}, as well as in novel settings with inconstant input sizes that we denote as multi-settings. All three networks are trained given the same regret budgets using our approach from Section \ref{sec:our_loss}. Additionally, we vary regret budgets and further investigate the differences between the architectures using novel validation procedures.  In all experiments, the valuations of all participants are additive and are independently drawn for each item from the Uniform distribution: $v_i (j) \in \mathbb{U}[0, 1]$. Training details, hyperparameters, and additional results are reported in the Appendix.

\subsection{Comparison of architectures under constant input sizes}\label{sec:experiments_uni}

The settings only differ in the number of participants $n$ and items $m$, so we denote them as $n$x$m$. We conduct experiments in five settings: \{1x2, 2x2, 2x3, 2x5, 3x10\}. The 1x2 setting is the celebrated Manelli-Vincent auction, the analytical solution for which is provided in \cite{manelli2006bundling}. The optimal revenue for this auction equals $0.55$. For the rest of the settings, the analytical solutions are unknown.

\begin{table}[t]
\centering
\caption{Architecture comparison. Like in \cite{dutting2019machine}, revenue is summed over participants and regret is averaged over participants. The highest revenue in a setting is highlighted with bold font. For brevity, we only report aggregated standard deviations: the average standard deviation of revenue is $0.006$ for 1x2, $0.011$ for 2x2, $0.009$ for 2x3, $0.033$ for 2x5, $0.019$ for 3x10; the average standard deviation of regret is $0.00018$ for $R_{max}=10^{-3}$, $0.00003$ for $R_{max}=10^{-4}$.}
\label{table:uni}
\begin{tabular}{llllllll}
\toprule
\multicolumn{1}{c}{\multirow{2}{*}{$R_{max}$}} & \multicolumn{1}{c}{\multirow{2}{*}{setting}} & \multicolumn{2}{c}{RegretNet} & \multicolumn{2}{c}{EquivariantNet} & \multicolumn{2}{c}{RegretFormer} \\
\multicolumn{1}{c}{}                           & \multicolumn{1}{c}{}                         & revenue       & regret        & revenue          & regret          & revenue        & regret         \\
\midrule
$10^{-3}$ & 1x2 & 0.572 & 0.0007 & \textbf{0.586} & 0.00065 & 0.571 & 0.00075 \\
 & 2x2 & 0.889 & 0.00055 & 0.878 & 0.0008 & \textbf{0.908} & 0.00054 \\
 & 2x3 & 1.317 & 0.00102 & 1.365 & 0.00084 & \textbf{1.416} & 0.00089 \\
 & 2x5 & 2.339 & 0.00142 & 2.437 & 0.00146 & \textbf{2.453} & 0.00102 \\
 & 3x10 & 5.59 & 0.00204 & 5.744 & 0.00167 & \textbf{6.121} & 0.00179 \\
\midrule
$10^{-4}$ & 1x2 & 0.551 & 0.00007 & 0.548 & 0.00013 & \textbf{0.556} & 0.00014 \\
 & 2x2 & 0.825 & 0.00005 & 0.75 & 0.00005 & \textbf{0.861} & 0.00006 \\
 & 2x3 & 1.249 & 0.00007 & 1.226 & 0.0001 & \textbf{1.327} & 0.00011 \\
 & 2x5 & 2.121 & 0.00013 & 2.168 & 0.00017 & \textbf{2.339} & 0.00015 \\
 & 3x10 & 5.02 & 0.00062 & 5.12 & 0.00025 & \textbf{5.745} & 0.00022 \\
\bottomrule
\toprule
\multicolumn{1}{c}{\multirow{2}{*}{$R_{max}$}} & \multicolumn{1}{c}{\multirow{2}{*}{setting}} & \multicolumn{2}{c}{VCG} & \multicolumn{2}{c}{Myerson item-wise} & \multicolumn{2}{c}{Myerson bundled} \\
\multicolumn{1}{c}{}                           & \multicolumn{1}{c}{}                         & revenue       & regret        & revenue          & regret          & revenue        & regret         \\
\midrule
--                                              & 1x2                                          & 0         & 0       & 0.5            & 0          & 0.544            & 0        \\
                                               & 2x2                                          & 0.667         & 0       & 0.833            & 0         & 0.839          & 0        \\
                                               & 2x3                                          & 1         & 0       & 1.25            & 0         & 1.278          & 0        \\
                                               & 2x5                                          & 1.667         & 0       & 2.083            & 0           & 2.188          & 0         \\
                                               & 3x10                                         & 5         & 0       & 5.312            & 0         & 5.003          & 0        \\
\bottomrule
\end{tabular}
\end{table}

\begin{table}[t]
\centering
\caption{Ratio of the estimated regret to the regret budget. Our approach from Section \ref{sec:our_loss} implies that this ratio should be close to 1 during training. Optimal misreports are estimated by minimizing (\ref{eq:loss_inner}) using 50 and 1000 gradient descent steps during training and validation, respectively.}
\label{table:ratio}
\begin{tabular}{llllllll}
\toprule
\multirow{2}{*}{$R_{max}$} & \multirow{2}{*}{setting} & \multicolumn{2}{c}{RegretNet} & \multicolumn{2}{c}{EquivariantNet} & \multicolumn{2}{c}{RegretFormer} \\
 &  & train & valid & train & valid & train & valid \\
\midrule
$10^{-3}$ & 1x2 & 1.12 & 1.22 & 1.04 & 1.11 & 1.01 & 1.31 \\
 & 2x2 & 0.97 & 1.24 & 1.41 & 1.82 & 0.89 & 1.19 \\
 & 2x3 & 1.07 & 1.55 & 1.11 & 1.23 & 1.02 & 1.26 \\
 & 2x5 & 0.94 & 1.21 & 1.11 & 1.2 & 0.8 & 0.83 \\
 & 3x10 & 0.89 & 1.09 & 0.9 & 0.87 & 1.03 & 0.88 \\
\midrule
$10^{-4}$ & 1x2 & 0.94 & 1.27 & 0.92 & 2.37 & 1.31 & 2.52 \\
 & 2x2 & 0.95 & 1.94 & 1.73 & 1.33 & 0.93 & 1.39 \\
 & 2x3 & 1.52 & 1.12 & 1.57 & 1.63 & 1.6 & 1.66 \\
 & 2x5 & 1.04 & 1.23 & 1.02 & 1.57 & 0.95 & 1.28 \\
 & 3x10 & 0.9 & 3.71 & 1.05 & 1.46 & 0.88 & 1.15 \\
\bottomrule
\end{tabular}
\end{table}

We compare our and existing neural architectures in the five settings given two different regret budgets $R_{max} \in \{10^{-3}, 10^{-4}\}$. We additionally include the classic VCG \cite{vickrey1961counterspeculation,clarke1971multipart,groves1973incentives} and Myerson \cite{myerson1981optimal} mechanisms for comparison, both of which are DSIC and ex-post IR. The Myerson auctions are optimal for $m=1$. We report its two extensions to multi-good settings: the `item-wise` where each item is sold separately and the `bundled` where all items are sold as a single bundle. 

We report the results in Table \ref{table:uni}. Additionally, we report the learning curves in the Appendix.

In all settings but 1x2 and given both regret budgets, RegretFormer outperforms both RegretNet and EquivariantNet in revenue. The performance gap is absent in the simplest setting but becomes especially prominent in the larger settings. While permutation-equivariance of RegretFormer likely plays a role, it cannot fully explain the results since EquivarintNet also has this property. We therefore attribute the superiority of RegretFormer to the expressivity of attention layers.

\begin{figure}[t]
    \centering
    
    \begin{minipage}{1\textwidth}
    \centering
    \begin{subfigure}{.245\textwidth}
        \centering
        \caption{}
        \includegraphics[width=\linewidth]{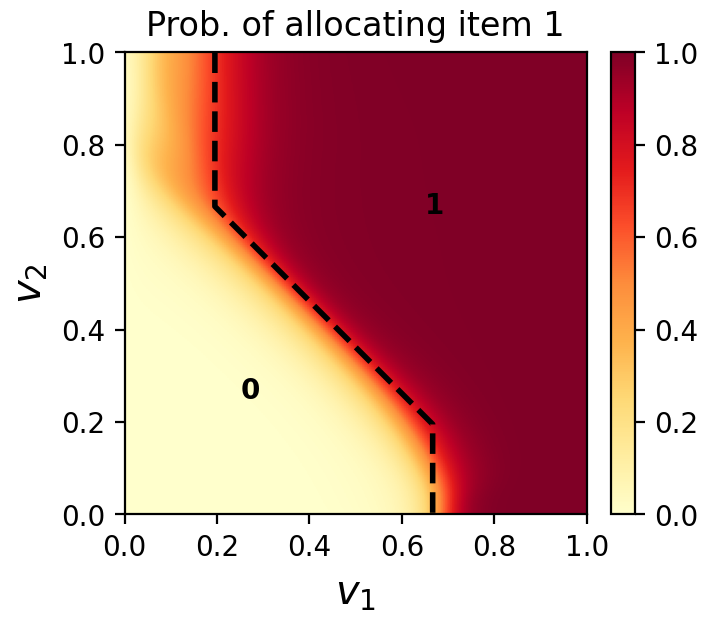}
    \end{subfigure}%
    \begin{subfigure}{.245\textwidth}
        \centering
        \caption{}
        \includegraphics[width=\linewidth]{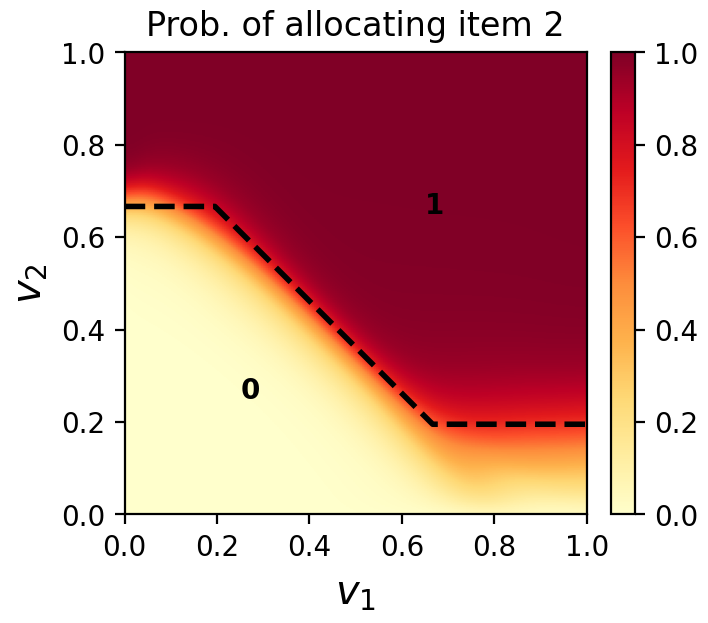}
    \end{subfigure}%
    \begin{subfigure}{.245\textwidth}
        \centering
        \caption{}
        \includegraphics[width=\linewidth]{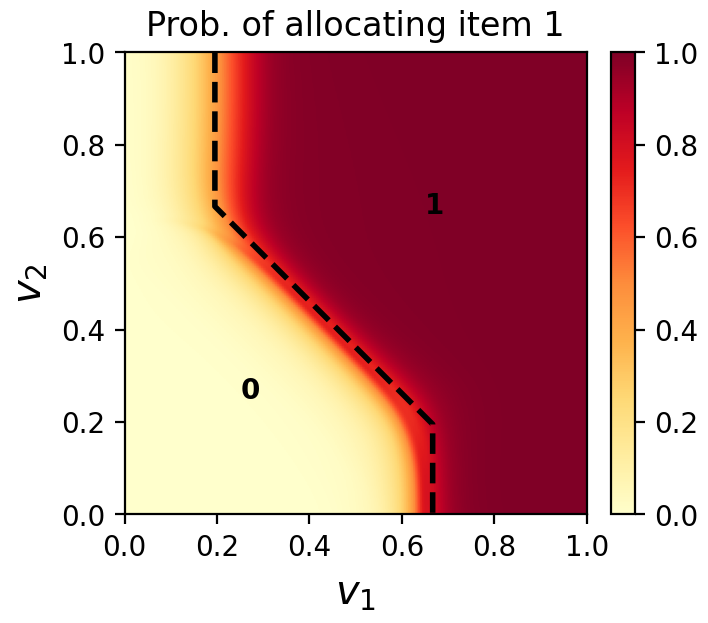}
    \end{subfigure}%
    \begin{subfigure}{.245\textwidth}
        \centering
        \caption{}
        \includegraphics[width=\linewidth]{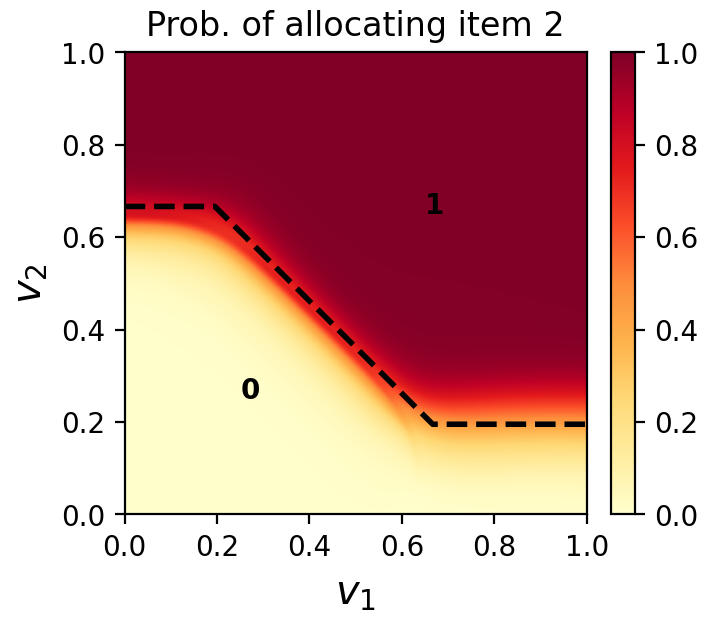}
    \end{subfigure}%
    \end{minipage}
    
    \begin{minipage}{1\textwidth}
    \begin{subfigure}{.245\textwidth}
        \centering
        \caption{}
        \includegraphics[width=\linewidth]{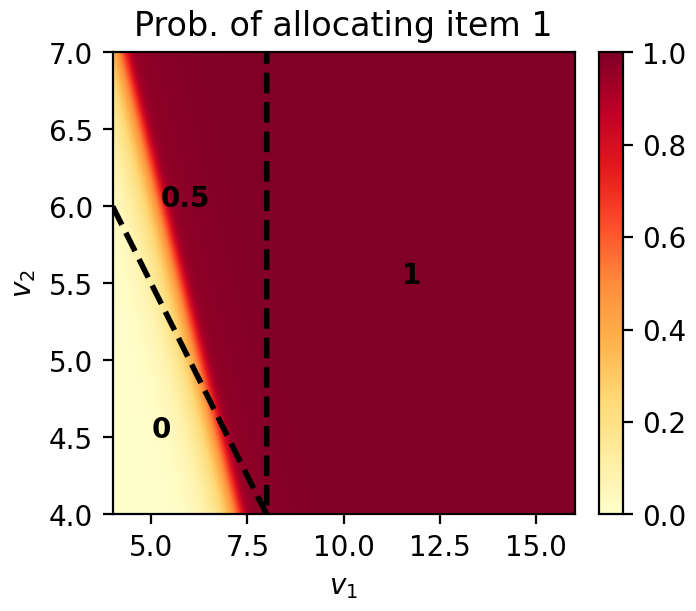}
    \end{subfigure}%
    \hfill%
    \begin{subfigure}{.245\textwidth}
        \centering
        \caption{}
        \includegraphics[width=\linewidth]{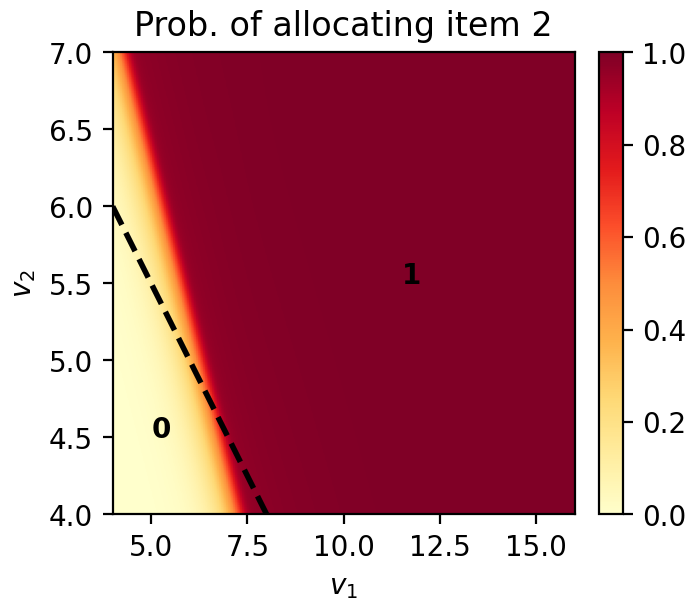}
    \end{subfigure}%
    \hfill%
    \begin{subfigure}{.245\textwidth}
        \centering
        \caption{}
        \includegraphics[width=\linewidth]{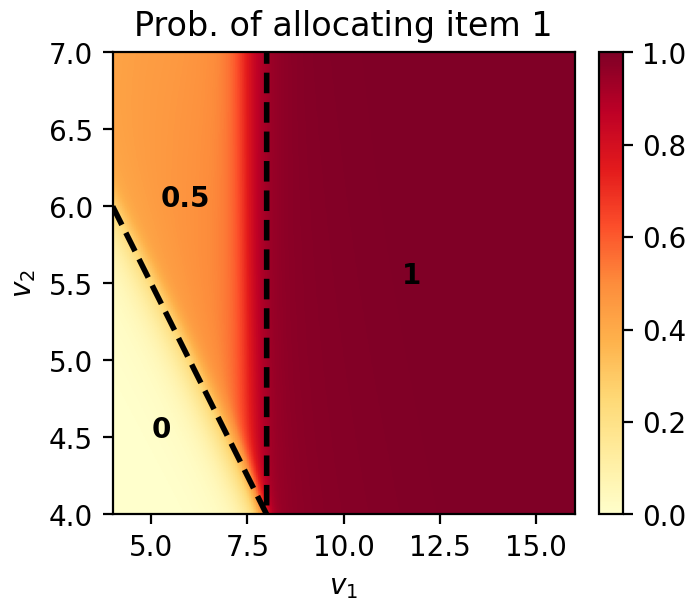}
    \end{subfigure}%
    \hfill%
    \begin{subfigure}{.245\textwidth}
        \centering
        \caption{}
        \includegraphics[width=\linewidth]{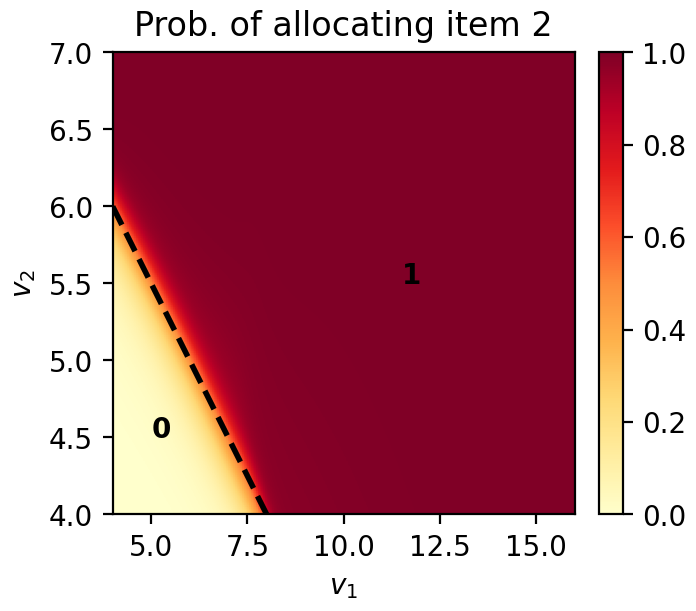}
    \end{subfigure}%
    \end{minipage}
    
    \caption{Allocation probabilities: (a, b) RegretNet in 1x2; (c, d) RegretFormer in 1x2; (e, f) RegretFormer without PE in the asymmetric setting; (g, h) RegretFormer with PE in the asymmetric setting. The optimal mechanisms are described by the regions separated by the dashed black lines, with the numbers in black denoting the optimal probability of allocation in the region.}
    \label{fig:alloc_probs}
\end{figure}

Additionally, in Figure \ref{fig:alloc_probs}, we provide solutions found by RegretNet and RegretFormer in the 1x2 setting. Both networks find allocation probabilities that smoothly approximate the optimal solution.

In the bigger settings, EquivariantNet can also outperform RegretNet. This result is somewhat surprising since it was not observed by the authors of EquivariantNet, but it can simply be explained by the fact that they do not test the architecture in settings that are complex enough.

Given the smallest regret budget, both baselines sometimes underperform Myerson. While RegretFormer outperforms Myerson in revenue, its regret is still non-zero. This highlights a potential problem of the regret-based approach: when any violations of DSIC are highly undesirable, a better revenue might be achieved with classic mechanisms that are also guaranteed to be DSIC.

A potential downside of RegretFormer compared to the baselines is that it takes longer to train and requires more parameters to reach optimal performance. We elaborate on this in the Appendix.

\subsection{Does regret budget function as indented?}

The core feature of our approach from Section \ref{sec:our_loss} is that the resulting regret should be uniquely and unambiguously specified through the regret budget. In particular, the total regret normalized by the total revenue should equal the regret budget. To verify this quantitatively, we present the ratios of the estimated normalized total regret to the regret budget in Table \ref{table:ratio}. A ratio being higher than 1 means that the regret budget is exceeded, and vice versa. Below we summarize our findings.

During training, the ratios are indeed close to 1 for all three architectures, meaning that our approach functions as intended. Additionally, we report the ratios estimated during validation, which involves more gradient descent steps to find optimal misreports than during training, making the regret estimates more precise. Unsurprisingly, these are usually than the ratios estimated during training. For a tighter regret approximation and a better match with the budget, one could increase the number of the inner optimization steps during training, provided that a longer training time is acceptable.

\subsection{Is the performance gap genuine?}\label{sec:experiments_val}

In Table \ref{table:uni}, we have observed that RegretFormer outperforms both baselines and have attributed its success to the expressivity of attention layers. However, there is an alternative explanation.

\begin{table}[t]
\centering
\caption{Cross-misreport regret estimates. The highest regret for a network is highlighted with bold.}
\label{table:cross_misreport}
\begin{tabular}{lllll}
\toprule
setting & regret of & \multicolumn{3}{c}{misreports of} \\
        &               & RegretNet    & EquivariantNet & RegretFormer \\
\midrule
1x2  & RegretNet & \textbf{0.00079} & 0.00043 & 0.00074 \\
 & EquivariantNet & 0.00102 & 0.00071 & \textbf{0.00129} \\
 & RegretFormer & 0.00076 & 0.00046 & \textbf{0.00092} \\
\midrule
2x2     & RegretNet & \textbf{0.00050} & 0.00031 & 0.00024 \\
 & EquivariantNet & 0.00021 & \textbf{0.00050} & 0.00022 \\
 & RegretFormer & \textbf{0.00065} & 0.00056 & 0.00059 \\
\midrule
2x3     & RegretNet & \textbf{0.00116} & 0.00062 & 0.00044 \\
 & EquivariantNet & 0.00020 & \textbf{0.00071} & 0.00028 \\
 & RegretFormer & 0.00072 & 0.00087 & \textbf{0.00094} \\
\midrule
2x5     & RegretNet & \textbf{0.00149} & 0.00065 & 0.00060 \\
 & EquivariantNet & 0.00075 & \textbf{0.00142} & 0.00071 \\
 & RegretFormer & 0.00089 & \textbf{0.00120} & 0.00109 \\
\midrule
3x10    & RegretNet & \textbf{0.00198} & 0.00035 & 0.00012 \\
 & EquivariantNet & 0.00035 & \textbf{0.00168} & 0.00014 \\
 & RegretFormer & 0.00178 & 0.00178 & \textbf{0.00222} \\
\bottomrule
\end{tabular}
\end{table}

Recall that regret is approximated in the inner optimization (\ref{eq:loss_inner}) by repeatedly back-propagating through the whole neural network $w$. It follows that regret approximation depends on the neural architecture, including the number, the types, the size, and the order of layers. Due to absence of a regret approximation procedure that is identical for all networks, there is no guarantee that similar regret estimates between different neural architectures correspond to actual similar regret values, i.e. for different architectures $w_1, w_2: \tilde{R}(w_1) \approx \tilde{R}(w_2) \centernot\implies R(w_1) \approx R(w_2)$, where $R(w)$ is the expected regret and $\tilde{R}(w)$ is its approximation. In particular, we are concerned whether RegretFormer achieves higher revenues due to approximating the regret worse, rather than due to maximizing the revenue better. We design two procedures to test this hypothesis.

The first procedure is denoted as `cross-misreports`. It is based on estimating the regret of one network on the optimal misreports approximated by another network: $\widetilde{R}_i^{cross} (w_1, w_2) = \frac{1}{\left|B\right|} \sum_{l \in B} \widetilde{rgt}_i(v_i^{\prime l}(w_2), v^l; w_1)$. We apply this procedure to the trained networks in all five settings given $R_{max} = 10^{-3}$. If all networks approximate regret equally well, we expect the regret estimated on cross-misreports to not exceed the normally computed regret. However, if the regret estimates of RegretFormer were higher on the misreports estimated by RegretNet or EquivariantNet, this would point towards poor regret approximation by RegretFormer.

We report the results in Table \ref{table:cross_misreport}. We do not find any evidence that RegretFormer underestimates the regret: in all settings, it finds misreports that produce regret either higher than or within a standard deviation from the misreports of the other two architectures.

The second procedure is based on network distillation. The results are similar to Table \ref{table:cross_misreport} in that RegretFormer does not appear to underestimate the regret, so we report them in the Appendix.

\subsection{Learning asymmetric mechanisms using positional encoding}\label{sec:exp_asym}

\begin{table}[t]
\centering
\caption{Asymmetric setting from \citet{daskalakis2017strong}, $R_{max} = 10^{-3}$}
\label{table:asymmetric}
\begin{tabular}{llllllll}
\toprule
\multicolumn{2}{l}{Optimal} & \multicolumn{2}{l}{RegretNet} & \multicolumn{2}{l}{RegretFormer} & \multicolumn{2}{l}{RegretFormer + PE} \\
\midrule
revenue & regret & revenue & regret & revenue & regret & revenue & regret \\
9.781 & 0 & 9.951 & 0.0072 & 9.967 & 0.0102 & 10.056 & 0.0099 \\
\bottomrule
\end{tabular}
\end{table}

To handle asymmetric distributions, our architecture can be augmented with Positional Encoding (PE) - a common technique for transformers to incorporate information about order of elements. To demonstrate this, we study a setting with one bidder, two items, and non-identically distributed valuations over items, where $v_1 \sim U[4, 16]$ and $v_2 \sim U[4,7]$. The optimal mechanism is given by \citet{daskalakis2017strong}. We apply RegretNet, RegretFormer without PE, and RegretFormer with PE given $R_{max} = 10^{-3}$. As PE, we use a common technique that adds arbitrary numbers from [-1, 1] to the input (estimated as sine and cosine functions of position-dependent frequencies) proposed by \citet{vaswani2017attention}. Note that this PE has no learnable parameters.

The results are reported in the Table \ref{table:asymmetric}. While RegretFormer with PE unsurprisingly performs on par with RegretNet, RegretFormer without PE learns a symmetric solution that is not much worse. In Figure \ref{fig:alloc_probs}, we provide illustrations of solutions found by the two versions of RegretFormer. These show that RegretFormer with PE closely approximates the allocation probabilities of the optimal mechanism, whereas RegretFormer without PE finds a symmetric solution that resembles some middle ground between the optimal allocation probabilities for the two items.

\subsection{Comparison of architectures in multi-settings}\label{sec:experiments_multi}

In practice, it may be desirable for a single network to be applicable to multiple input sizes, e.g. to save computation or due to limited data. To test our network in such cases, we introduce \emph{multi-setting}. A multi-setting is a uniform mixture of constant-sized settings studied so far. In our experiments, we mix the following settings: \{2x3, 2x4, 2x5, 2x6, 2x7, 3x3, 3x4, 3x5, 3x6, 3x7\}. We set $R_{max}=10^{-3}$. To adapt RegretNet to multi-settings, we fix the input size for the maximum possible number of items and participants in a multi-setting and pad unused valuations with zeros.



\begin{table}[t]
\centering
\caption{Multi-setting results. We report the highest regret identified with the cross-misreport procedure (see full cross-misreport results in the Appendix). (*) Note that EquivariantNet achieves unrealistically high revenue for a much higher regret.}
\label{table:multi_big}
\begin{tabular}{lllllll}
\toprule
\multirow{2}{*}{Setting} & \multicolumn{2}{c}{RegretNet} &  \multicolumn{2}{c}{EquivariantNet$^{(*)}$}  &  \multicolumn{2}{c}{RegretFormer} \\
              & revenue        & regret      & revenue        & regret       & revenue         & regret        \\
\midrule
average                  & 2.573         & 0.00352       & 2.889            & 0.00989         & 2.703 & 0.00391 \\
\midrule
2x3 & 1.386            & 0.00305         & 1.504            & 0.00554         & 1.432            & 0.00246         \\
2x4 & 1.855            & 0.00341         & 2.070            & 0.00925         & 1.951            & 0.00317         \\
2x5 & 2.339            & 0.00362         & 2.646            & 0.01270         & 2.471            & 0.00391         \\
2x6 & 2.866            & 0.00425         & 3.226            & 0.01597         & 3.006            & 0.00439         \\
2x7 & 3.346            & 0.00457         & 3.834            & 0.01951         & 3.553            & 0.00481         \\
3x3 & 1.652            & 0.00322         & 1.795            & 0.00358         & 1.708            & 0.00251         \\
3x4 & 2.217            & 0.00264         & 2.449            & 0.00508         & 2.312            & 0.00336         \\
3x5 & 2.786            & 0.00277         & 3.108            & 0.00709         & 2.911            & 0.00421         \\
3x6 & 3.362            & 0.00340         & 3.787            & 0.00916         & 3.521            & 0.00476         \\
3x7 & 3.921            & 0.00430         & 4.467            & 0.01101         & 4.140            & 0.00553         \\
\bottomrule
\end{tabular}
\end{table}

The results are presented in Table \ref{table:multi_big}. Comparing RegretFormer with RegretNet, our architecture consistently achieves higher revenue for the same regret budget. While EquivariantNet achieves even higher revenue, after applying our cross-misreport validation procedure we find that EquivariantNet fails to minimize regret due to poorly approximating optimal misreports. Full results of cross-misreport validation, as well as out-of-setting experiments, are reported in the Appendix.

\section{Conclusion}

In this study, we achieve new state-of-the-art in the application of deep learning to optimal auction design. Our RegretFormer leverages the recent advances in deep learning to unlock the full potential of the regret-based optimization while enforcing the equivariance and the invariance to the order of participants and items. We test the effectiveness of RegretFormer in multiple experimental settings and find that our network consistently outperforms the existing analogues. In addition, we rethink the objective formulation of RegretNet. The resulting loss function disentangles balancing the revenue-regret trade-off and optimizing the performance, as well as reduces the burden of hyperparameter tuning. Finally, we suggest two validation procedures for comparing regret-based approaches that may find use in future studies.


\section*{Acknowledgments}

\textit{All authors}. We sincerely thank Xeniya Adayeva for creating the illustration of RegretFormer \ref{fig:scheme}. You can contact Xeniya at \textit{xeniya.adayeva@gmail.com} regarding scientific illustrations.

\textit{Dmitry Ivanov}. This research was supported in part through computational resources of HPC facilities at HSE University, Russian Federation. Support from the Basic Research Program of the National Research University Higher School of Economics is gratefully acknowledged.

\bibliographystyle{abbrvnat}
\bibliography{main}

\section*{Checklist}

\begin{enumerate}

\item For all authors...
\begin{enumerate}
  \item Do the main claims made in the abstract and introduction accurately reflect the paper's contributions and scope?
    \answerYes{}
  \item Did you describe the limitations of your work?
    \answerYes{} \textbf{See the end of Section 4.1.}
  \item Did you discuss any potential negative societal impacts of your work?
    \answerNo{} \textbf{We do not foresee such impact.}
  \item Have you read the ethics review guidelines and ensured that your paper conforms to them?
    \answerYes{}
\end{enumerate}

\item If you ran experiments...
\begin{enumerate}
  \item Did you include the code, data, and instructions needed to reproduce the main experimental results (either in the supplemental material or as a URL)?
    \answerYes{} \textbf{We include the code as a URL.}
  \item Did you specify all the training details (e.g., data splits, hyperparameters, how they were chosen)?
    \answerYes{} \textbf{See the Appendix.}
        \item Did you report error bars (e.g., with respect to the random seed after running experiments multiple times)?
    \answerYes{} \textbf{In Table 1, we report aggregated standard deviations. In the Appendix, we report the learning curves with the error bars.}
        \item Did you include the total amount of compute and the type of resources used (e.g., type of GPUs, internal cluster, or cloud provider)?
    \answerYes{} \textbf{See the Appendix.}
\end{enumerate}

\end{enumerate}
\newpage

\appendix
\section{Layers and architecture of RegretFormer}

\paragraph{Multi-head attention}

Popularized by \citet{vaswani2017attention}, the attention function maps a query vector and a set of key-value vector pairs to an output vector. The procedure is typically applied to a set or a sequence of queries. The output vector is a weighted sum of the values, and each weight reflects the compatibility of the query with the corresponding key. While different attention mechanisms exist, the softmax attention is the most common:


\begin{equation}
    \text{Attention}(Q, K, V) = \text{softmax}(\frac{QK^T}{\sqrt{d_k}})V
\end{equation}

\noindent where \(Q\), \(K\), and \(V\) are respectively the matrices of queries, keys, and values, and \(d_k\) is the number of keys. Self-attention is a special case of attention in which \(Q\), \(K\), and \(V\) are linear projections of the same inputs. Typically, layer normalization is applied to the input before projecting \cite{ba2016layer}. \citet{vaswani2017attention} also propose multi-head attention (MHA). In this extension, \emph{H} different attention heads are created, and for each attention head the input matrices are projected with head-specific weight matrices \(QW_{h}^{Q}\), \(KW_{h}^{K}\), \(VW_{h}^{V}\) to calculate the inputs to attention:

\begin{equation}
    \text{MHA}(Q, K, V) = \text{Concat}(head_{1}, ..., head_{H})W^{O}
\end{equation}
\begin{equation}
    head_{h} = \text{Attention}(QW_{h}^Q, KW_{h}^K, VW_{h}^V)
\end{equation}

Attention is equivariant to the order of elements in the input, which is a useful property when learning symmetric auctions since any optimal symmetric auction can be represented by a permutation-equivariant function \cite{daskalakis2012symmetries}. In applications where this order is important (e.g. order of words in a sentence) Positional Encoding (PE) is usually applied. This technique augments the initial representation of the input data with the information about the order of the elements. We demonstrate how PE can be applied to learn optimal asymmetric auctions in Section \ref{sec:exp_asym}.

\paragraph{Exchangeable Layers}

The exchangeable layer \cite{hartford2018deep} is inspired by deep sets \cite{zaheer2017deep} and is defined as follows. A layer is specified by the number of input channels $K$, the number of output channels $O$ and five learnable parameters $w_1, w_2, w_3, w_4 \in \mathbb{R}^{K\text{x}O}$ and $w_5 \in \mathbb{R}^O$. The input is a tensor \textit{B} of size \textit{(K, n, m)} and the output is tensor \textit{Y} of size \textit{(O, n, m)}. The element (i, j) of the \textit{o}-th output channel $Y_{i,j}^{(o)}$ is given by:

\begin{equation}\label{eq:exchangeable}
    Y_{i,j}^{(o)} = \sigma \left( \sum_{k=1}^K (w_{1}^{(k,o)} B_{i, j}^{(k)} + \frac{w_2^{(k,o)}}{n} \sum_{i'}B_{i', j}^{(k)} + \frac{w_3^{(k,o)}}{m} \sum_{j'}B_{i, j'}^{(k)} + \frac{w_4^{(k,o)}}{nm} \sum_{i',j'}B_{i', j'}^{(k)}) + w_5^{(o)} \right)
\end{equation}

This layer constitutes the main building block of EquivariantNet \cite{rahme2021permutation} and is also used as the first layer of RegretFormer.

\paragraph{Architecture of RegretFormer}

There are several high-level differences between the architectures of RegretFormer and RegretNet. Whereas RegretNet uses two separate networks to calculate allocations and payments, our architecture has a single shared network with both outputs. Unlike RegretNet where the input is flattened into a vector, the input of RegretFormer is the two-dimensional matrix of bids \(B^{nm}\). Furthermore, unlike RegretNet, \(n\) and \(m\) are not fixed in RegretFormer.

We now describe the architecture in detail. Note that each described layer except the output layers is followed by a Tanh activation.

First, we apply an exchangeable layer (\ref{eq:exchangeable}) to transform each bid into an initial vector of features that already contains information about other bids. According to our definition of the exchangeable layer, this requires adding third dummy dimension to the bid matrix \(B^{nm} \rightarrow B^{nm1}\):

\begin{equation}
    L_{1}^{nmk}=\text{ExchangeableLayer}(B^{nm1})
\end{equation}

Then, we sequentially apply several attention-based blocks. Each block consists of two multi-head self-attention layers with residual connections, one applied item-wise and one applied participant-wise. For each layer, we accordingly reshape the input. After applying the attention layers, we concatenate their predictions and apply the same fully-connected layer (FC) to the feature vectors of each bid (to reduce the dimensionality of the feature vectors to the initial size):

\begin{equation}
    L_{t+1,item}^{nmk}= \text{MHA}_{item}(L_{t}, L_{t}, L_{t})+L_{t}
\end{equation}
\begin{equation}
    L_{t+1,part}^{nmk}=\text{MHA}_{part}(L_{t}, L_{t}, L_{t})+L_{t}
\end{equation}
\begin{equation}
    L_{t+1}^{nmk}=\text{FC}_{t+1}(\text{Concat}(L_{t+1,item}, L_{t+1,part}))+L_{t}
\end{equation}

After applying the attention-based block $N$ times (in our experiments, we set $N$ to 1 or 2), we obtain the attended feature matrix \(L_{N+1}^{nmk}\). From this matrix, we create two separate matrices by averaging over one of the dimensions: the participant feature matrix \(P_{N+1}^{nk}=\frac{1}{m} \sum_j (L_{N+1}^{njk})\) and the item feature matrix \(I_{N+1}^{mk}=\frac{1}{n} \sum_i (L_{N+1}^{imk})\). These matrices are essentially embeddings of participants and items respectively and are used to compute the allocation matrix and the payment vector.

To compute the allocation matrix, we multiply the item and the participant matrices, which gives us an $n$ by $m$ matrix of unscaled probabilities (logits). Before scaling, we need to additionally consider the possibility of each item remaining unallocated. To this end, we introduce a dummy participant $n+1$, the unscaled probability for which is estimated for each item as a negated sum of the unscaled probabilities over the real participants. Finally, we apply the softmax function along the participants to scale the probabilities. When summarized, the allocation matrix is obtained the following way:

\begin{equation}
    L_{N+2}^{nm}=\text{MatMul}(P_{N+1}^{nk}, (I_{N+1}^{mk})^{T})
\end{equation}
\begin{equation}
    L_{norm}^{(n+1)m}=\text{Concat}(L^{nm}_{N+2}, -\sum_i (L_{N+2}^{im}))
\end{equation}
\begin{equation}
    Z_{out}^{(n+1)m}=\text{SoftMax}(L_{norm}^{(n+1)m})
\end{equation}

To estimate the payment vector, we average the participant feature matrix over the feature dimension and apply the sigmoid activation to scale the output between 0 and 1:

\begin{equation}
    \hat{P}_{out}^{n} = \text{Sigmoid}(\frac{1}{k} \sum_z P_{N+1}^{nz})
\end{equation}

\noindent Like in RegretNet, we calculate the final payments as \(p_i=\hat{p_i}\stackrel{~}\sum_{j=1}^mz_{ij}b_{ij}\) for \(i=1, \dots, n\).

\section{Technical details and hyperparameters}\label{sec:app_hyper}

In all experiments, all networks are trained for 200000 iterations of outer optimization, each iteration corresponding to one step of the optimizer on one mini-batch. The training dataset consists of 640000 profiles (same as in the RegretNet paper) divided into 1250 mini-batches of 512 profiles. The validation dataset consists of 4096 profiles divided into 128 batches of 32 profiles. The number of inner optimization steps per one outer update equals 50 during training and 1000 during validation. The learning rate equals $0.001$ for the outer optimization and $0.1$ for the inner optimization. Both use separate Adam optimizers \cite{kingma2015adam}. The hyperparameters related to the neural architectures are reported in Table \ref{tab:hyperparameters}. We report the sizes of neural networks in Table \ref{table:size}. All experiments are repeated three times and the average metrics are reported. Experiments are run on an internal cluster with V100 GPUs.

The hyperparameters related to our budget-based approach are the following. In all our experiments, we initialize $\gamma = 1$, set $\gamma_\Delta = 0.5$, set $R_{max}^{start} = 0.01$, and set such $R_{max}^{mult}$ that $R_{max}$ converges to $R_{max}^{end}$ in two-thirds of the training time. We set $R_{max}^{end} = 0.001$ by default but additionally investigate the effect of choosing a lower budget $R_{max}^{end} = 0.0001$.

\begin{table}[t]
\centering
\caption{Neural architecture hyperparameters}
\begin{tabular}{llllllll}
\toprule
Hyperparameter          & 1x2 & 2x2 & 2x3 & 2x5 & 3x10 & multi \\
\midrule
\textbf{RegretNet}      &     &     &     &     &      &           \\
fully-connected layers       & 3   & 3   & 3   & 6   & 6        & 6   \\
hidden dim                 & 200 & 200 & 200 & 200 & 200   & 200 \\
\midrule
\textbf{EquivariantNet} &     &     &     &     &      &           \\
exchangeable layers             & 3   & 3   & 5   & 6   & 6        & 6   \\
hidden dim                 & 32  & 32  & 32  & 32  & 32      & 32  \\
\midrule
\textbf{RegretFormer}             &     &     &     &          &       &     \\
exchangeable layers             & 1   & 1   & 1   & 1   & 1        & 1   \\
attention layers              & 1   & 1   & 1   & 2   & 2        & 2   \\
attention heads               & 2   & 2   & 2   & 4   & 4       & 4   \\
hidden dim                 & 32  & 64  & 64  & 128 & 128    & 128    \\
\bottomrule
\end{tabular}
\label{tab:hyperparameters}
\end{table}

\section{Additional results}\label{app:experiments}

\subsection{Learning curves}\label{app:experiments_curves}

\begin{figure}[h]
    \centering   
    \noindent\begin{minipage}{0.06\textwidth}
    \end{minipage}%
    \hfill%
    \begin{minipage}{0.94\textwidth}
    \centering
    \begin{subfigure}{.65\linewidth}
        \centering
        \includegraphics[width=\linewidth]{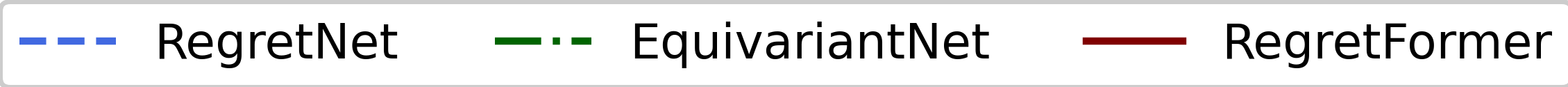}
    \end{subfigure}
    \end{minipage}
    
    \noindent\begin{minipage}{0.06\textwidth}
    $R_{max}$\\
    $10^{-3}$
    \end{minipage}%
    \hfill%
    \begin{minipage}{0.94\textwidth}
    \begin{subfigure}{.33\textwidth}
        \centering
        \caption{revenue}
        \includegraphics[width=\linewidth]{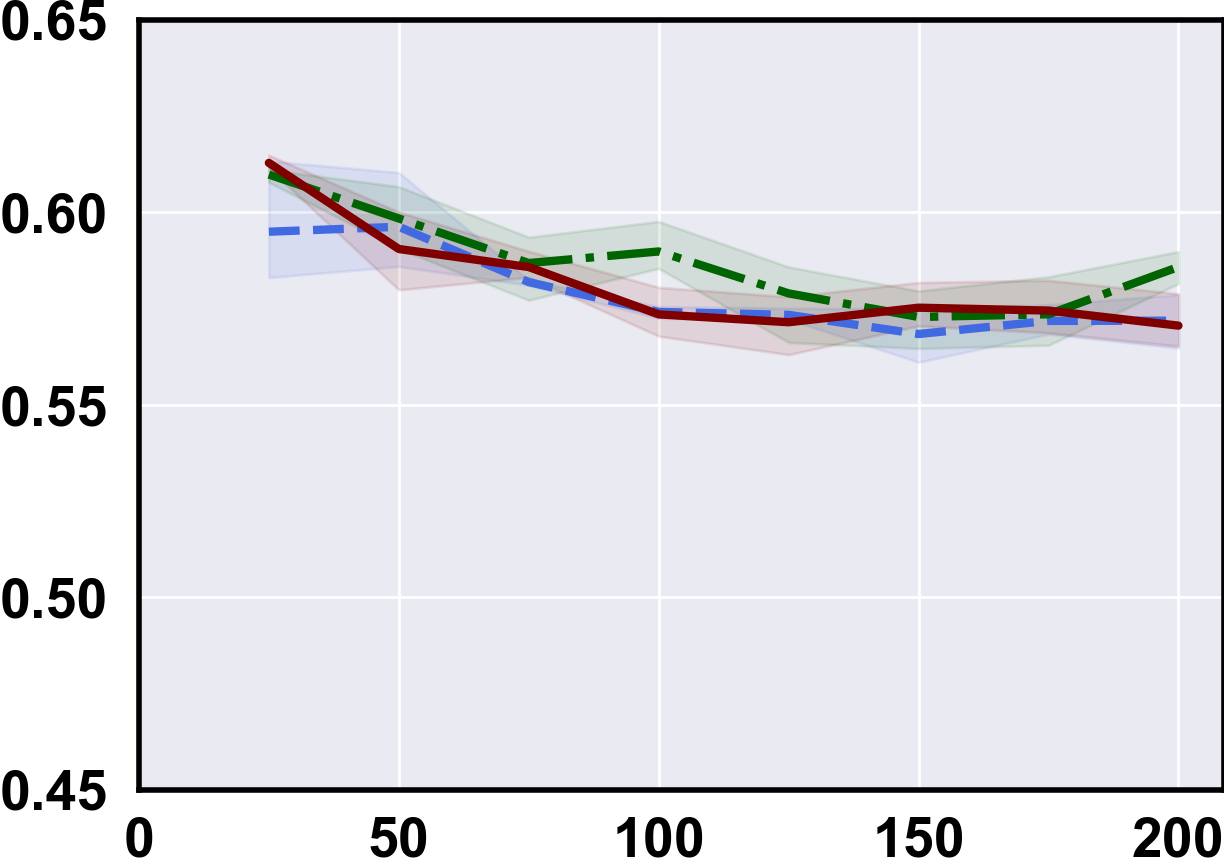}
    \end{subfigure}%
    \hfill%
    \begin{subfigure}{.33\textwidth}
        \centering
        \caption{regret}
        \includegraphics[width=\linewidth]{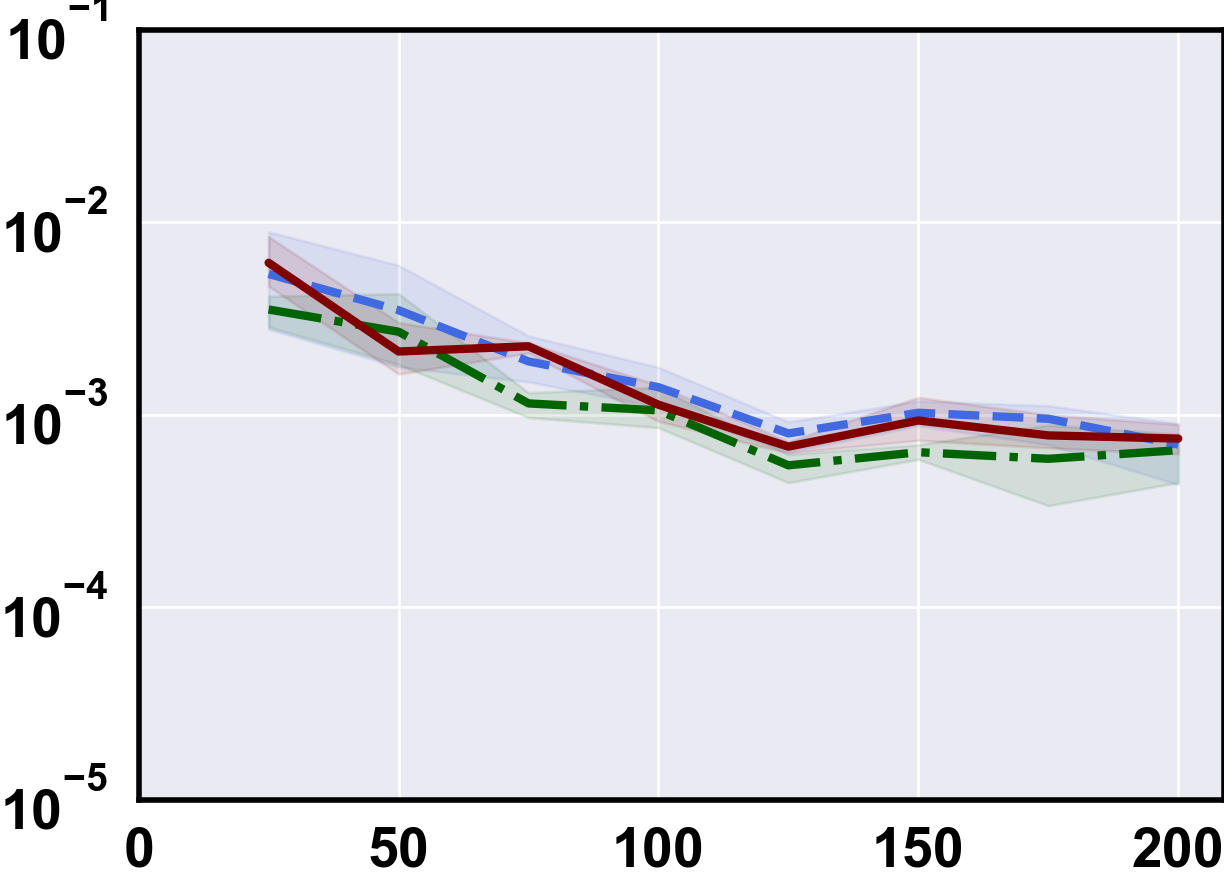}
    \end{subfigure}%
    \hfill%
    \begin{subfigure}{.32\textwidth}
        \centering
        \caption{penalty coefficient $\gamma$}
        \includegraphics[width=\linewidth]{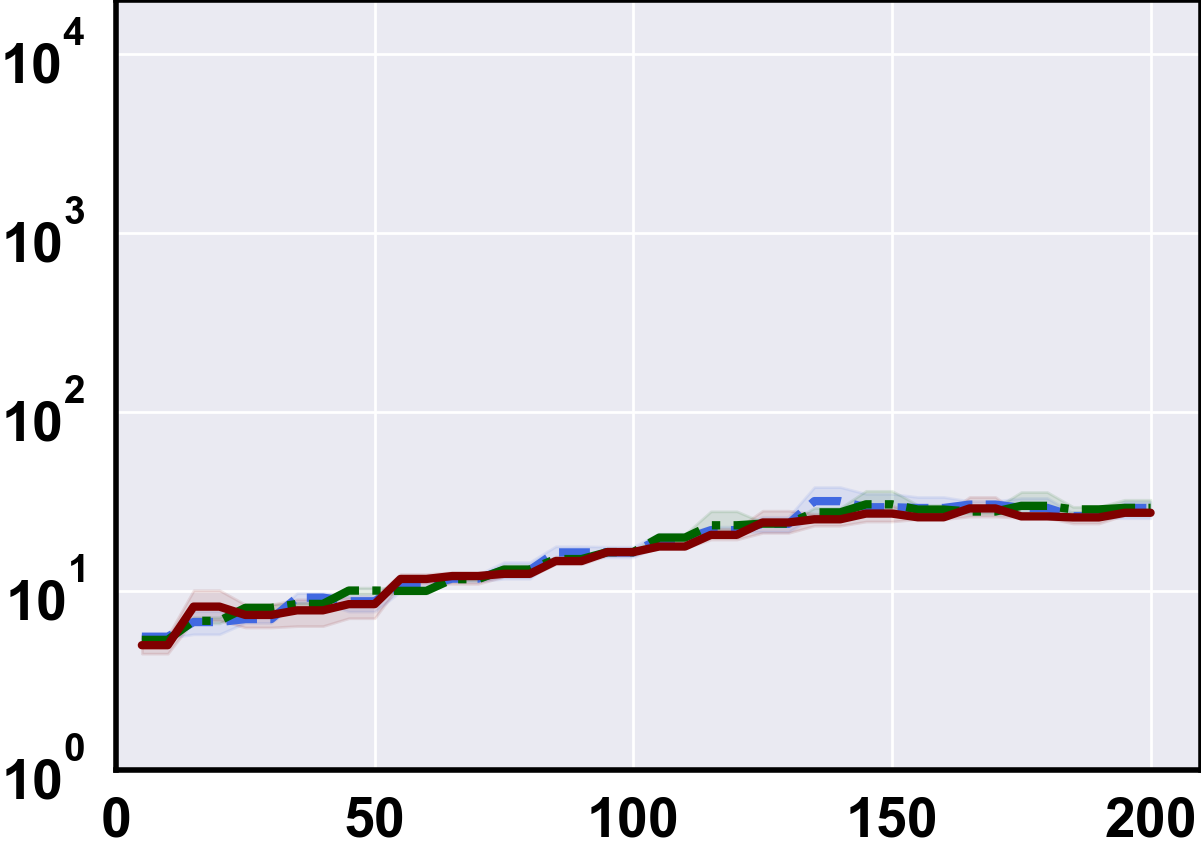}
    \end{subfigure}
    \end{minipage}

    \noindent\begin{minipage}{0.06\textwidth}
    $R_{max}$\\
    $10^{-4}$
    \end{minipage}%
    \hfill%
    \begin{minipage}{0.94\textwidth}
    \begin{subfigure}{.33\linewidth}
        \centering
        \includegraphics[width=\linewidth]{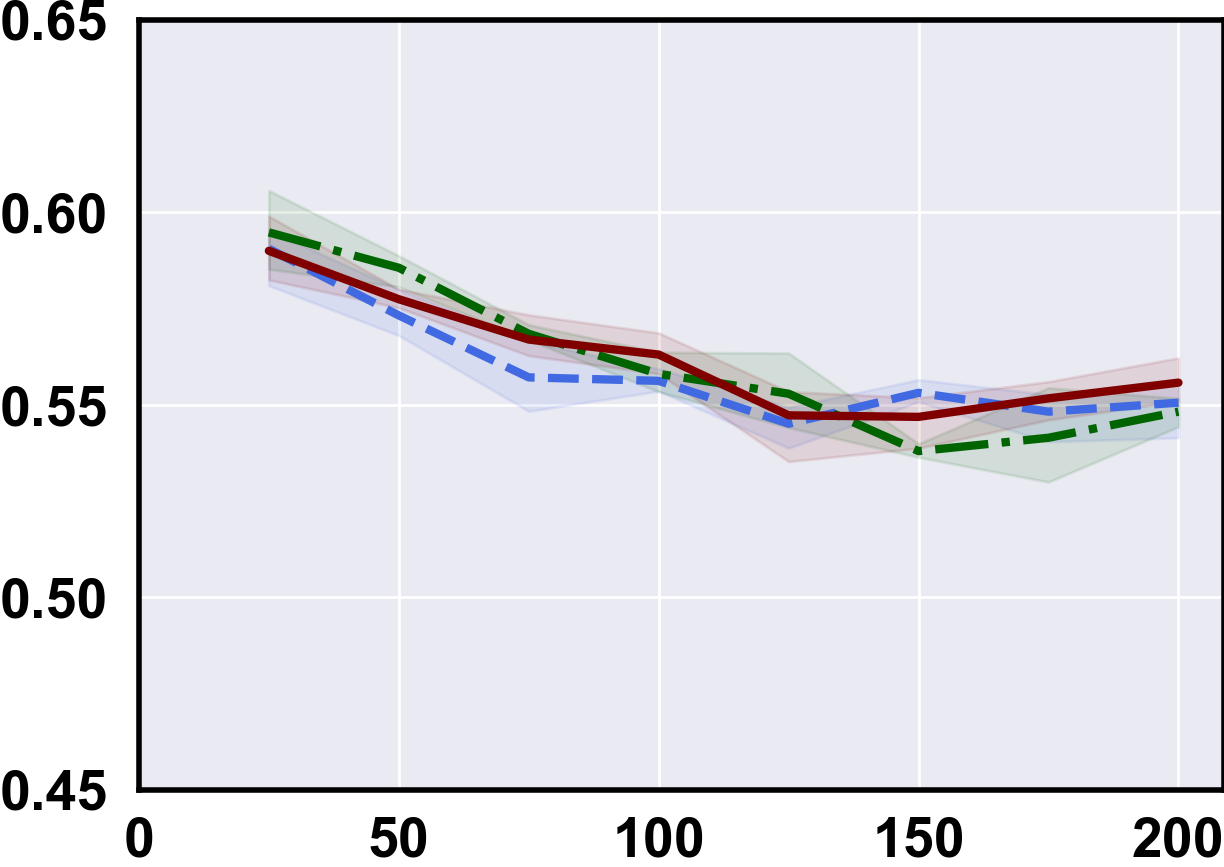}
    \end{subfigure}%
    \hfill%
    \begin{subfigure}{.33\linewidth}
        \centering
        \includegraphics[width=\linewidth]{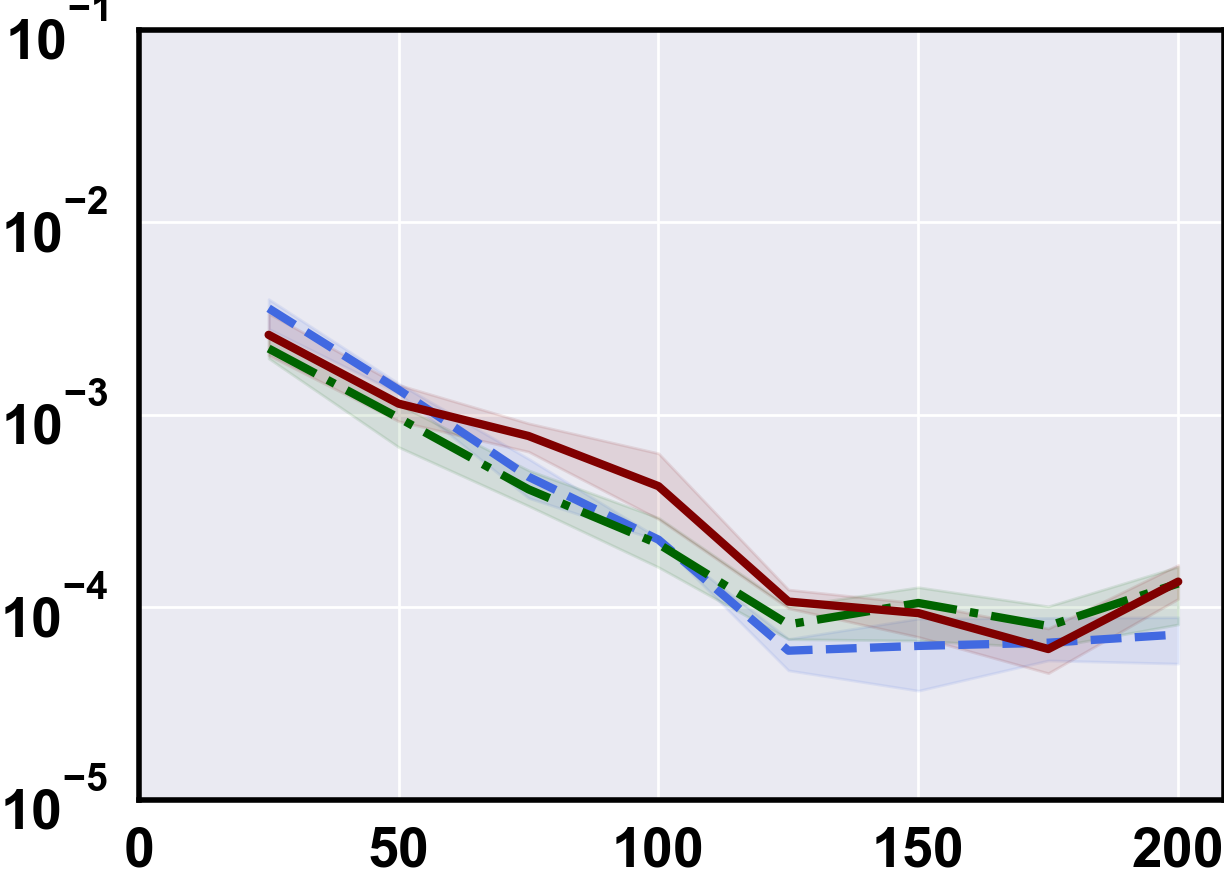}
    \end{subfigure}%
    \hfill%
    \begin{subfigure}{.32\linewidth}
        \centering
        \includegraphics[width=\linewidth]{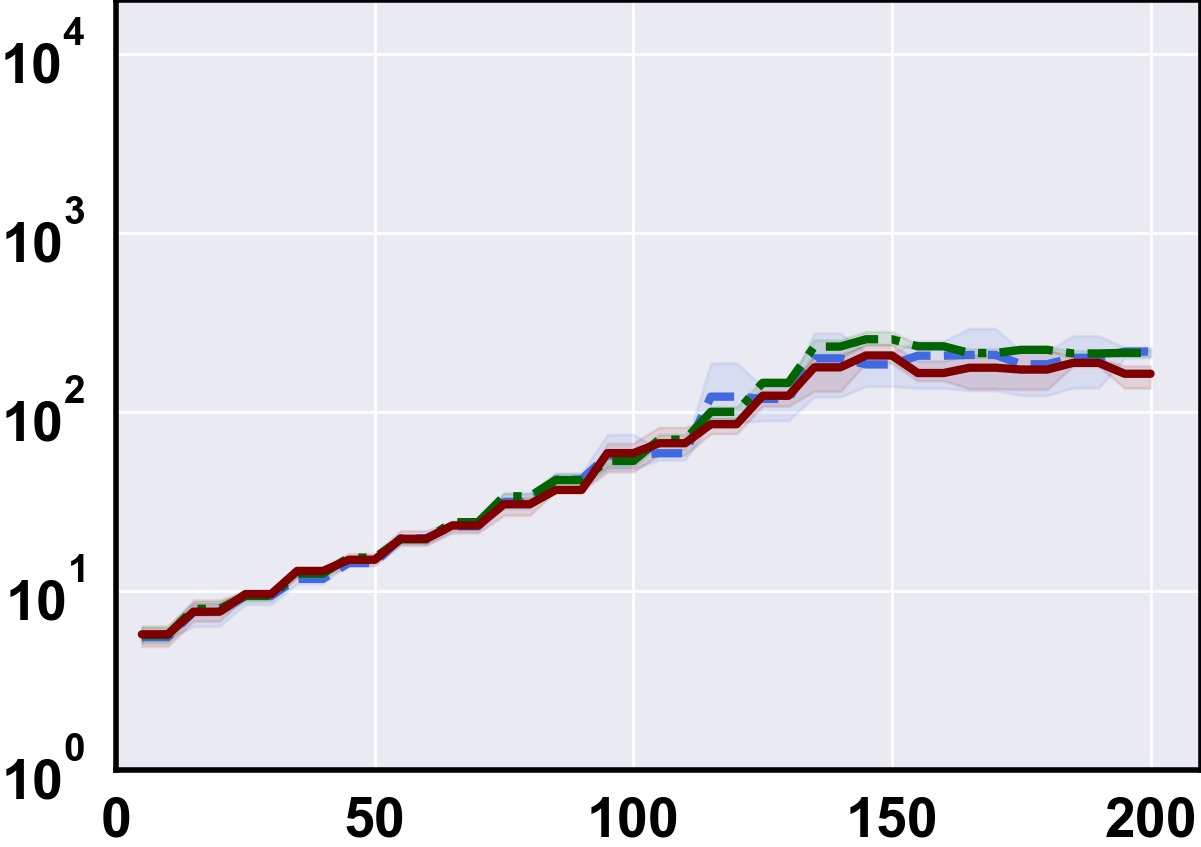}
    \end{subfigure}
    \end{minipage}

    \caption{Learning curves in the setting 1x2. The X-axis is in the thousands of training iterations. The shaded regions correspond to the min-max spread over three random seeds.}
    \label{fig:uni_1x2}
\end{figure}

\begin{figure}[t]
    \centering   
    \noindent\begin{minipage}{0.06\textwidth}
    \end{minipage}%
    \hfill%
    \begin{minipage}{0.94\textwidth}
    \centering
    \begin{subfigure}{.65\linewidth}
        \centering
        \includegraphics[width=\linewidth]{pics/legend.png}
    \end{subfigure}
    \end{minipage}
    
    \noindent\begin{minipage}{0.06\textwidth}
    $R_{max}$\\
    $10^{-3}$
    \end{minipage}%
    \hfill%
    \begin{minipage}{0.94\textwidth}
    \begin{subfigure}{.33\textwidth}
        \centering
        \caption{revenue}
        \includegraphics[width=\linewidth]{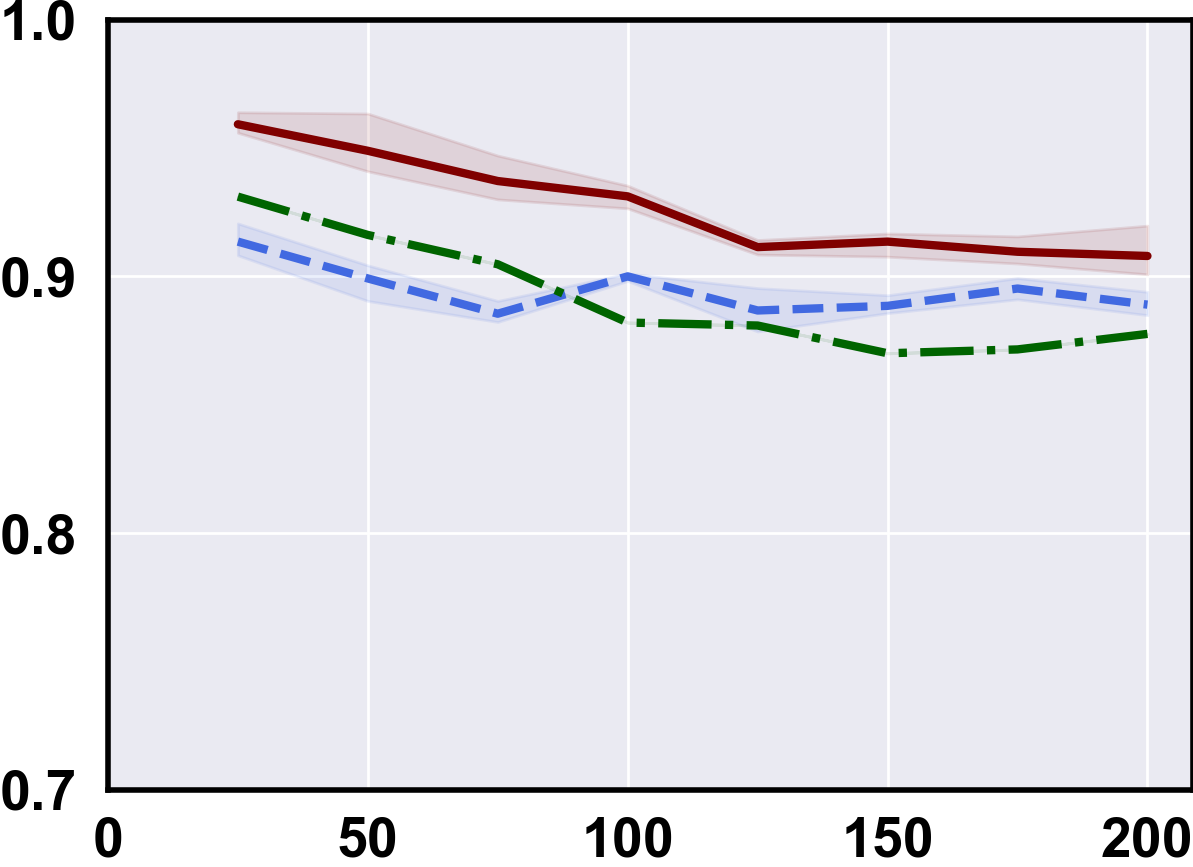}
    \end{subfigure}%
    \hfill%
    \begin{subfigure}{.33\textwidth}
        \centering
        \caption{regret}
        \includegraphics[width=\linewidth]{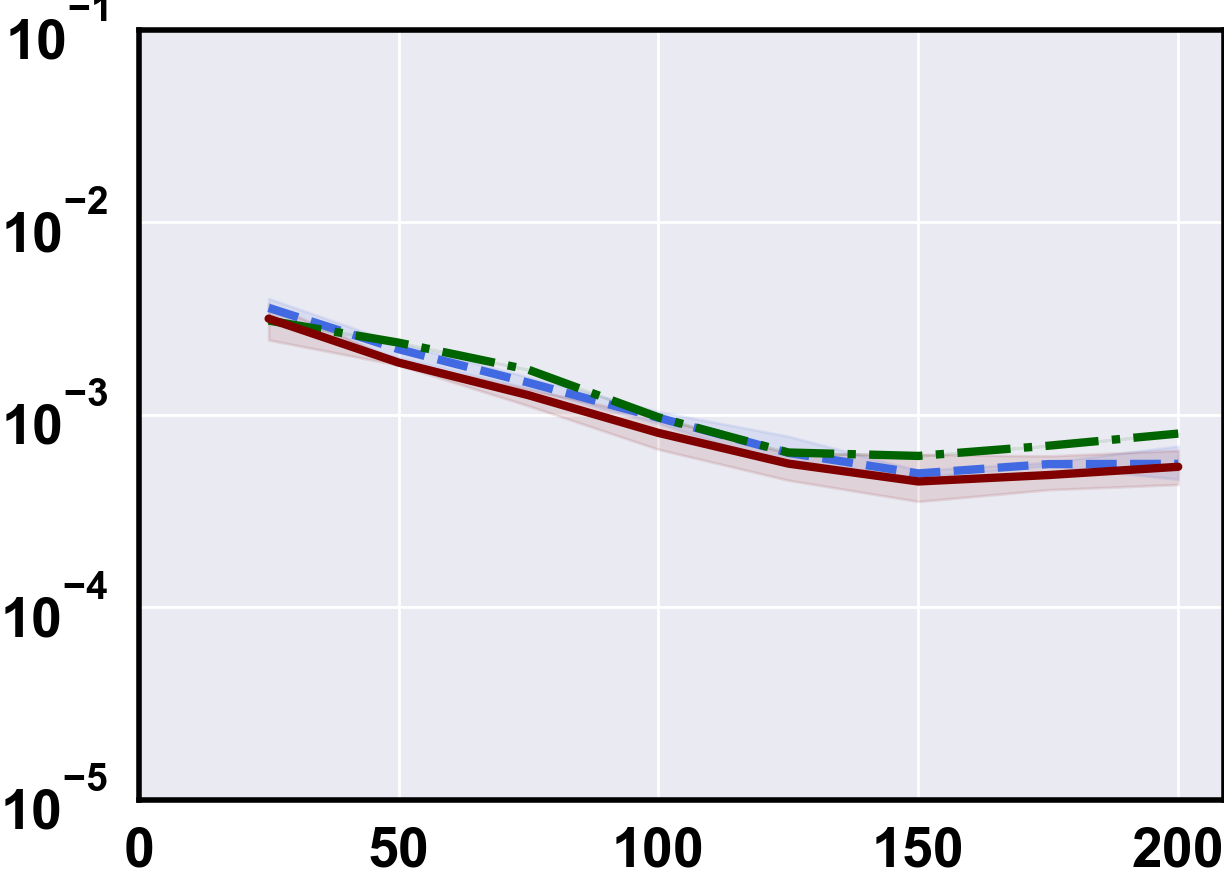}
    \end{subfigure}%
    \hfill%
    \begin{subfigure}{.32\textwidth}
        \centering
        \caption{penalty coefficient $\gamma$}
        \includegraphics[width=\linewidth]{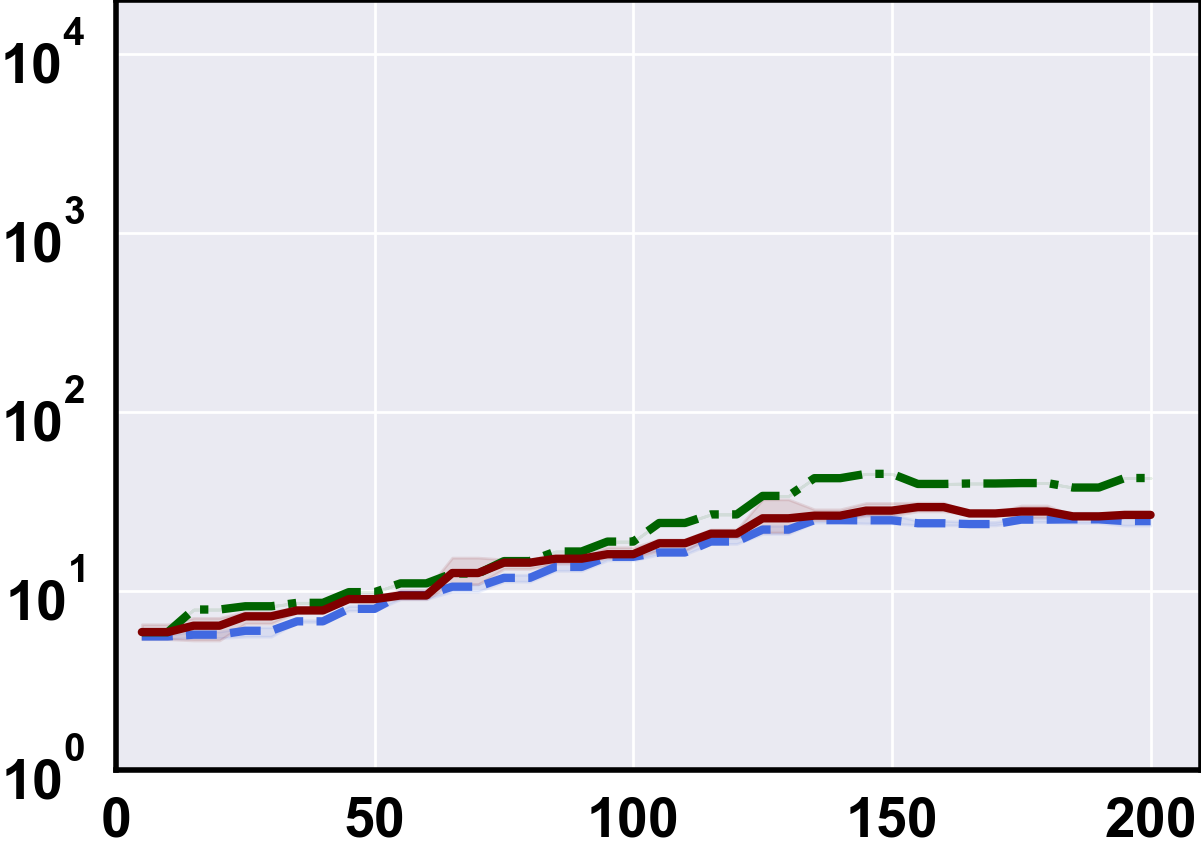}
    \end{subfigure}
    \end{minipage}

    \noindent\begin{minipage}{0.06\textwidth}
    $R_{max}$\\
    $10^{-4}$
    \end{minipage}%
    \hfill%
    \begin{minipage}{0.94\textwidth}
    \begin{subfigure}{.33\linewidth}
        \centering
        \includegraphics[width=\linewidth]{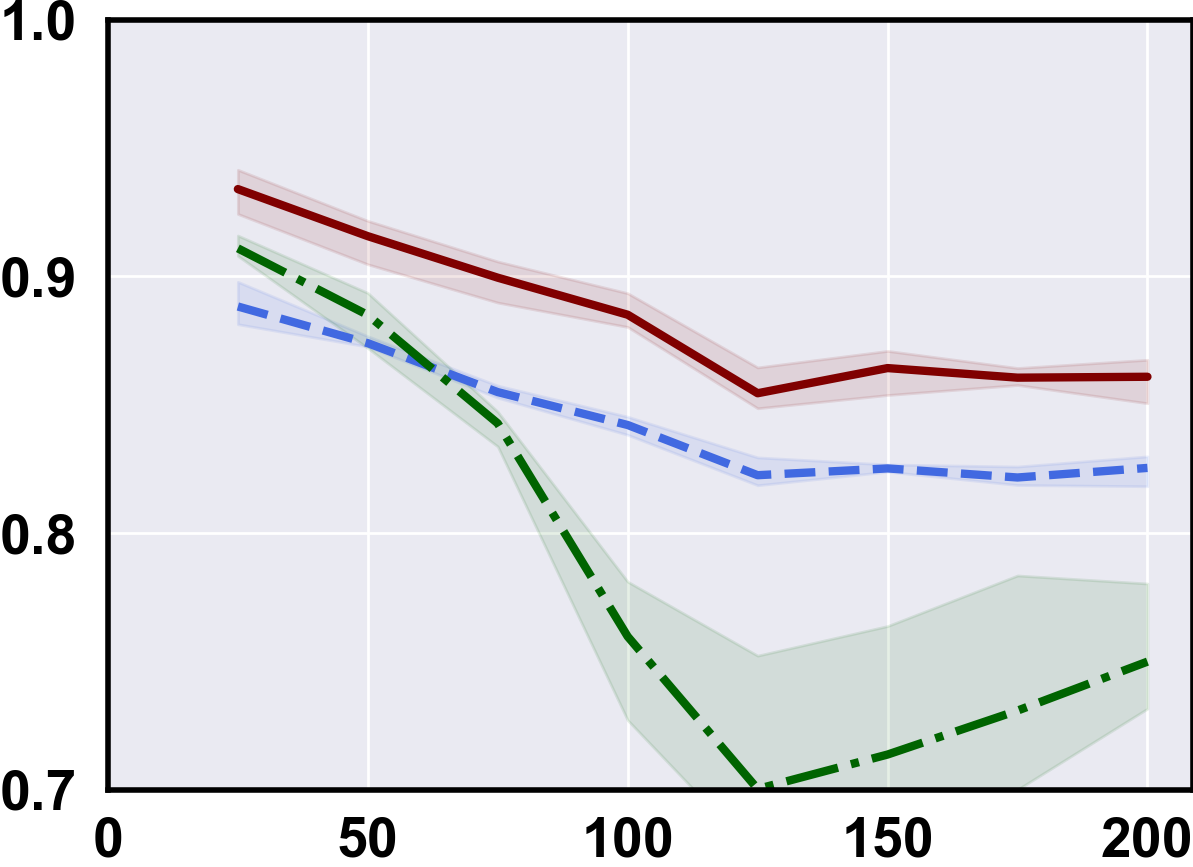}
    \end{subfigure}%
    \hfill%
    \begin{subfigure}{.33\linewidth}
        \centering
        \includegraphics[width=\linewidth]{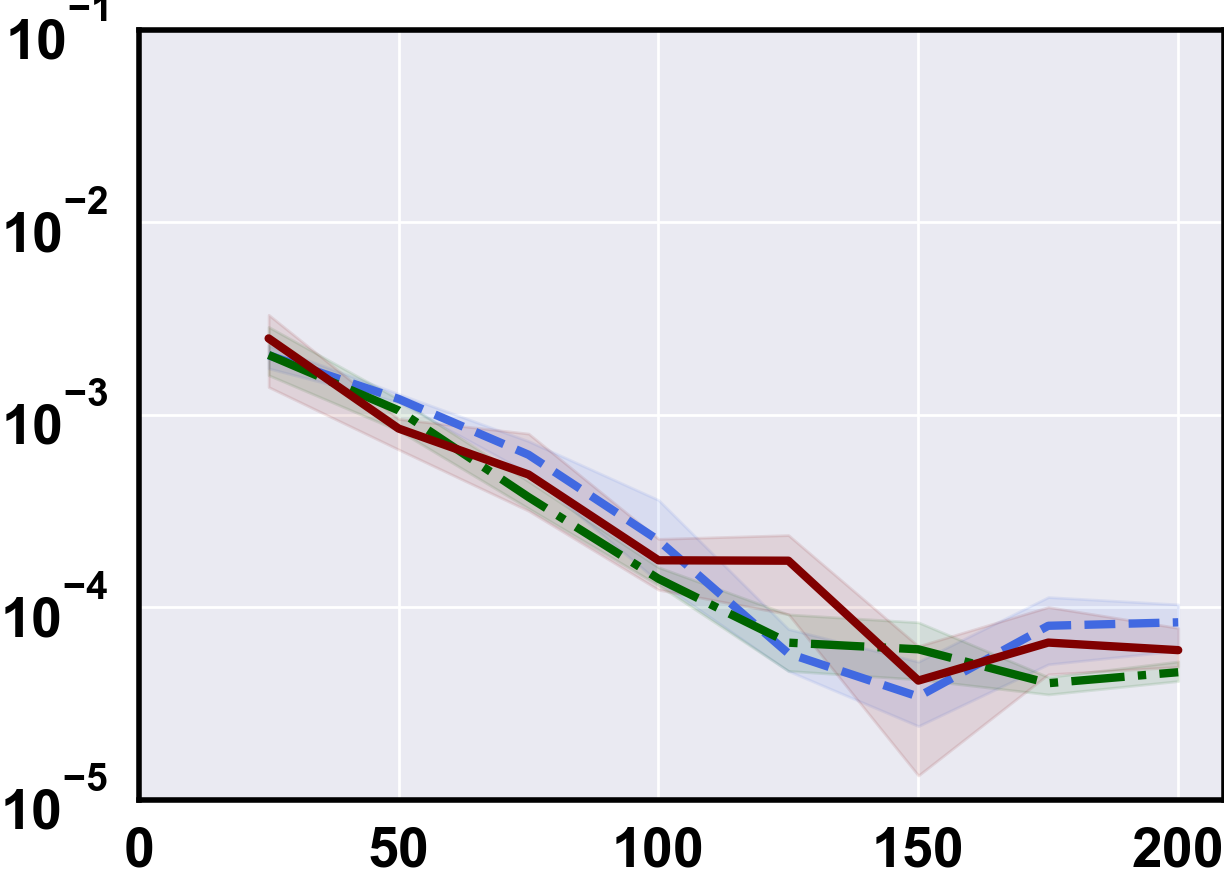}
    \end{subfigure}%
    \hfill%
    \begin{subfigure}{.32\linewidth}
        \centering
        \includegraphics[width=\linewidth]{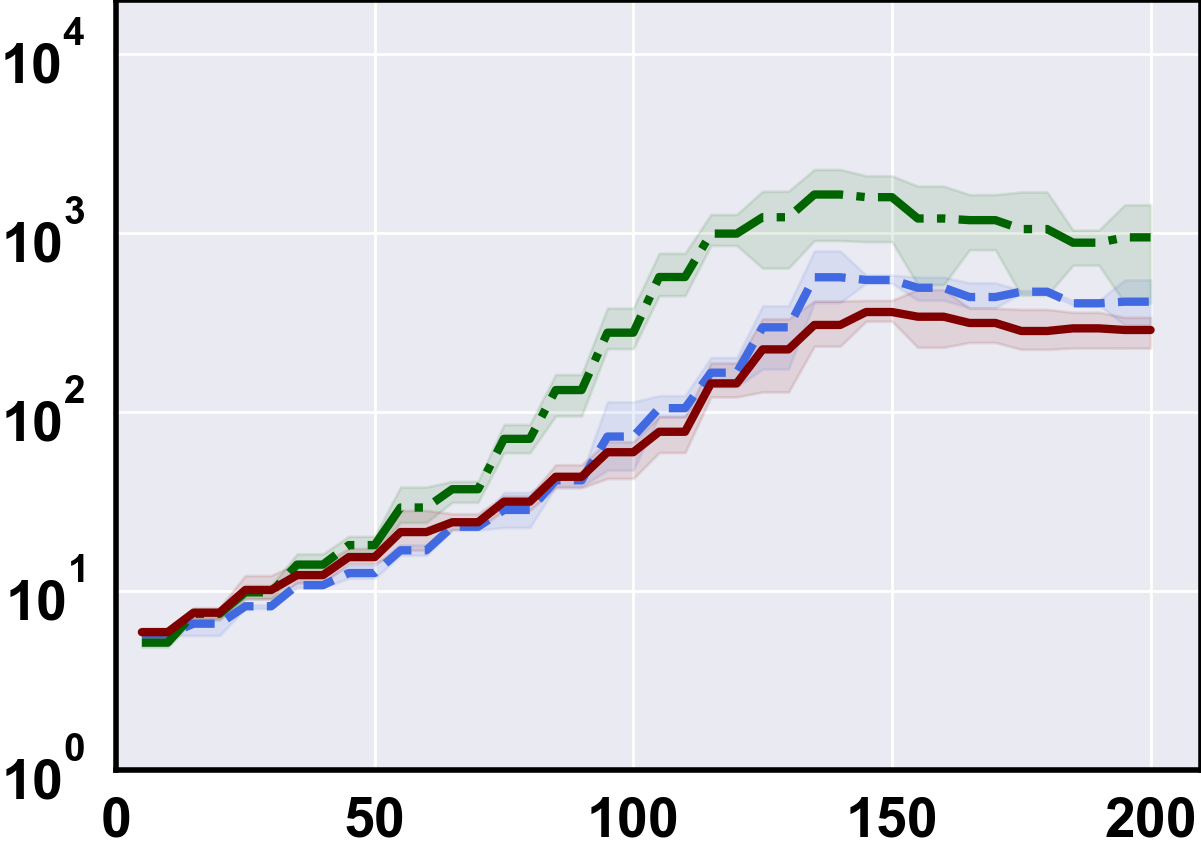}
    \end{subfigure}
    \end{minipage}
    
    \caption{Learning curves in the setting 2x2. The X-axis is in the thousands of training iterations. The shaded regions correspond to the min-max spread over three random seeds.}
    \label{fig:uni_2x2}
\end{figure}

\begin{figure}[t]
    \centering   
    \noindent\begin{minipage}{0.06\textwidth}
    \end{minipage}%
    \hfill%
    \begin{minipage}{0.94\textwidth}
    \centering
    \begin{subfigure}{.65\linewidth}
        \centering
        \includegraphics[width=\linewidth]{pics/legend.png}
    \end{subfigure}
    \end{minipage}
    
    \noindent\begin{minipage}{0.06\textwidth}
    $R_{max}$\\
    $10^{-3}$
    \end{minipage}%
    \hfill%
    \begin{minipage}{0.94\textwidth}
    \begin{subfigure}{.33\textwidth}
        \centering
        \caption{revenue}
        \includegraphics[width=\linewidth]{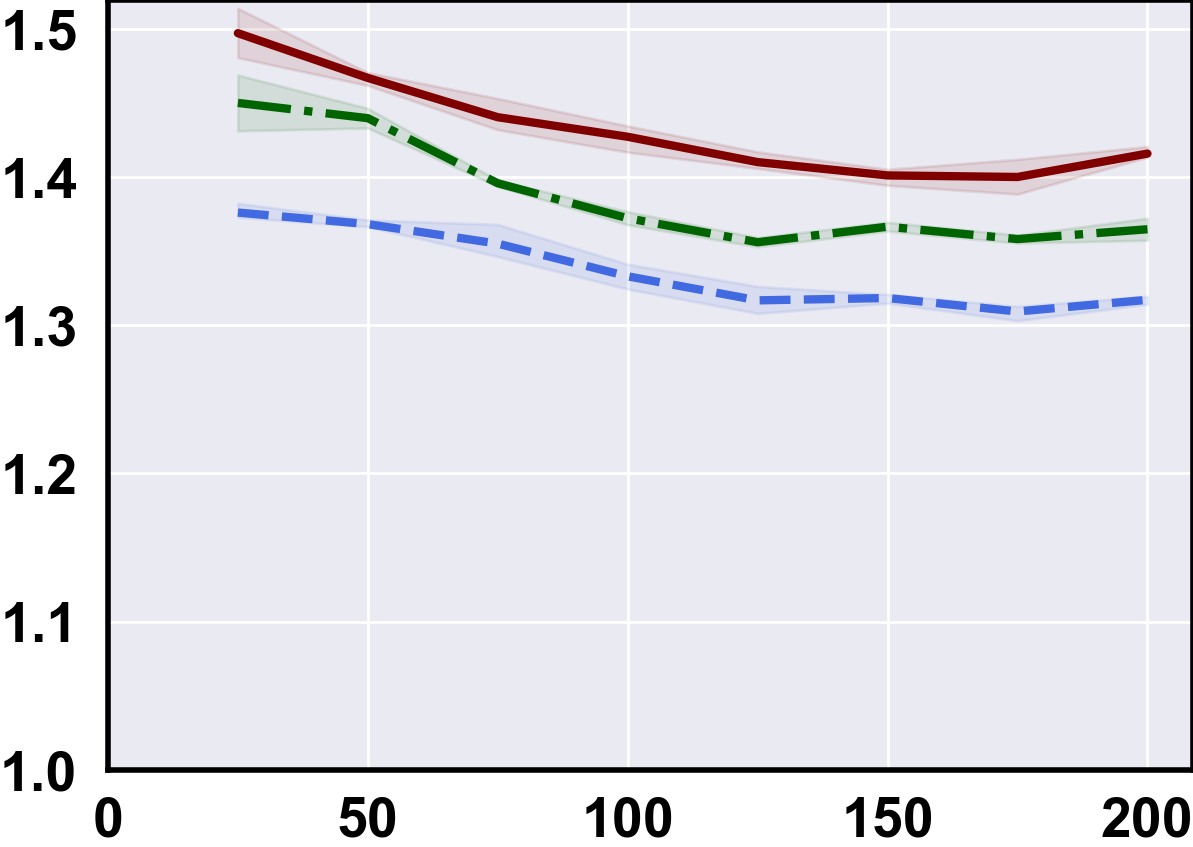}
    \end{subfigure}%
    \hfill%
    \begin{subfigure}{.33\textwidth}
        \centering
        \caption{regret}
        \includegraphics[width=\linewidth]{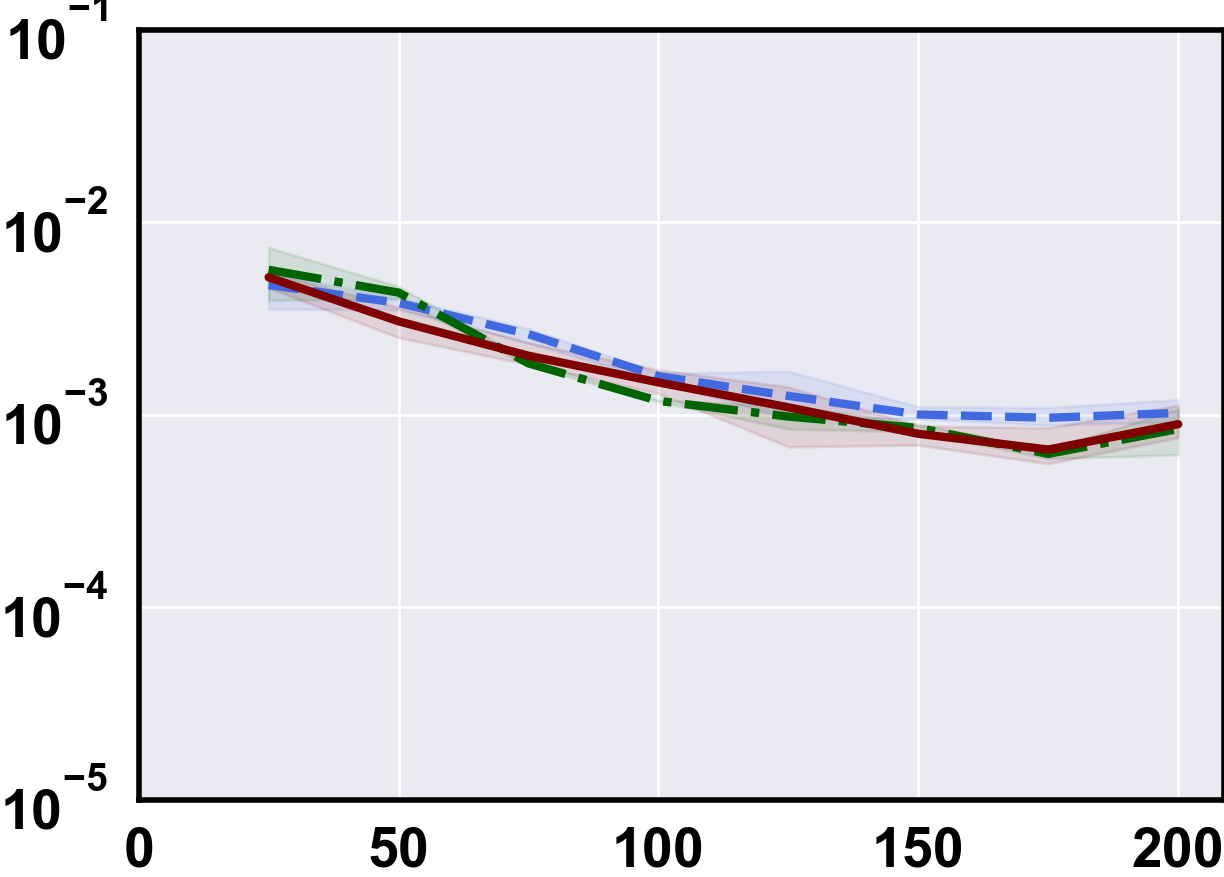}
    \end{subfigure}%
    \hfill%
    \begin{subfigure}{.32\textwidth}
        \centering
        \caption{penalty coefficient $\gamma$}
        \includegraphics[width=\linewidth]{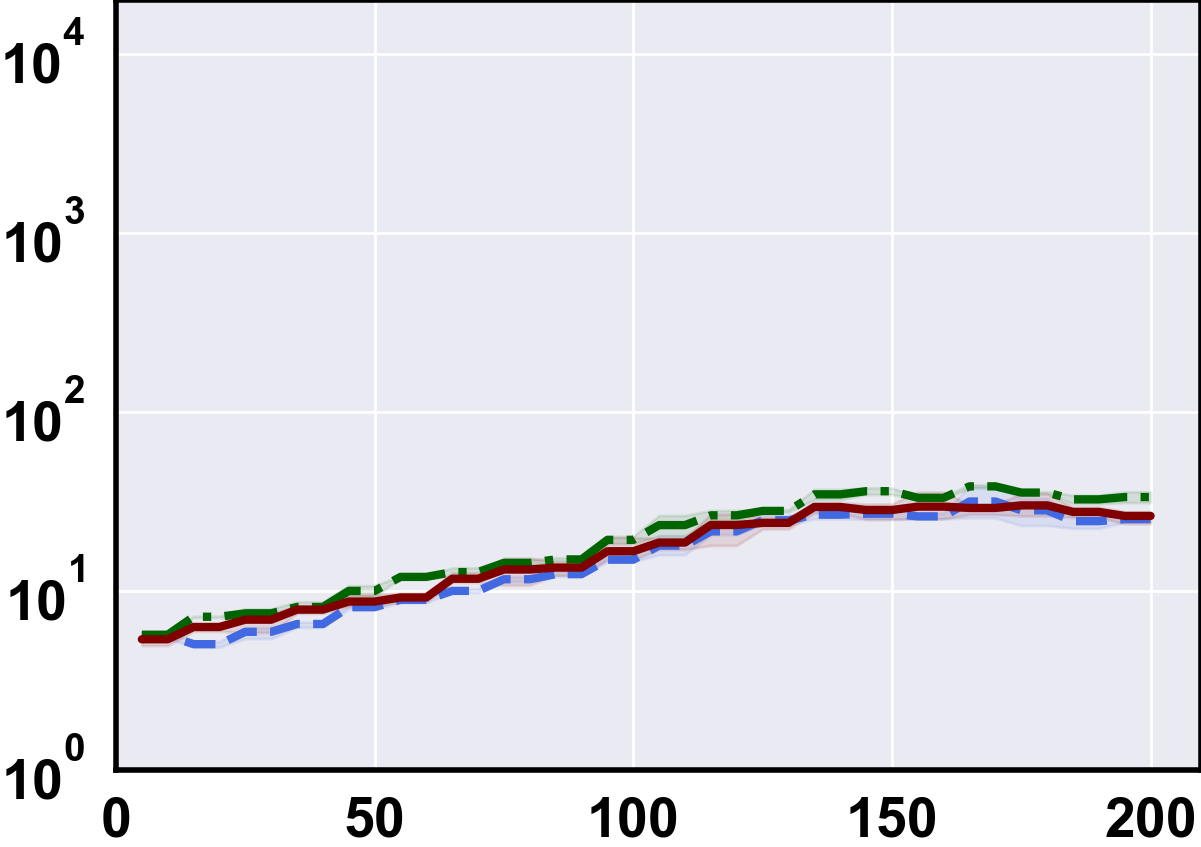}
    \end{subfigure}
    \end{minipage}

    \noindent\begin{minipage}{0.06\textwidth}
    $R_{max}$\\
    $10^{-4}$
    \end{minipage}%
    \hfill%
    \begin{minipage}{0.94\textwidth}
    \begin{subfigure}{.33\linewidth}
        \centering
        \includegraphics[width=\linewidth]{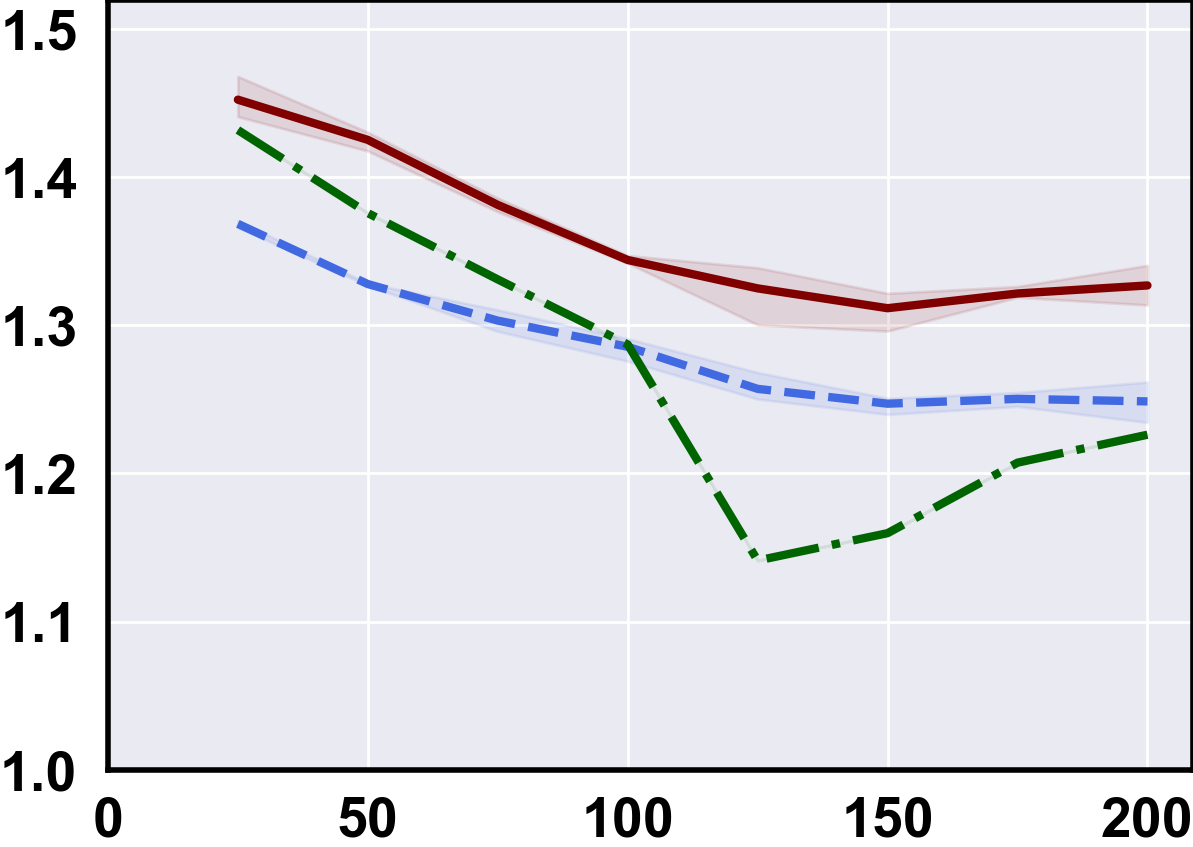}
    \end{subfigure}%
    \hfill%
    \begin{subfigure}{.33\linewidth}
        \centering
        \includegraphics[width=\linewidth]{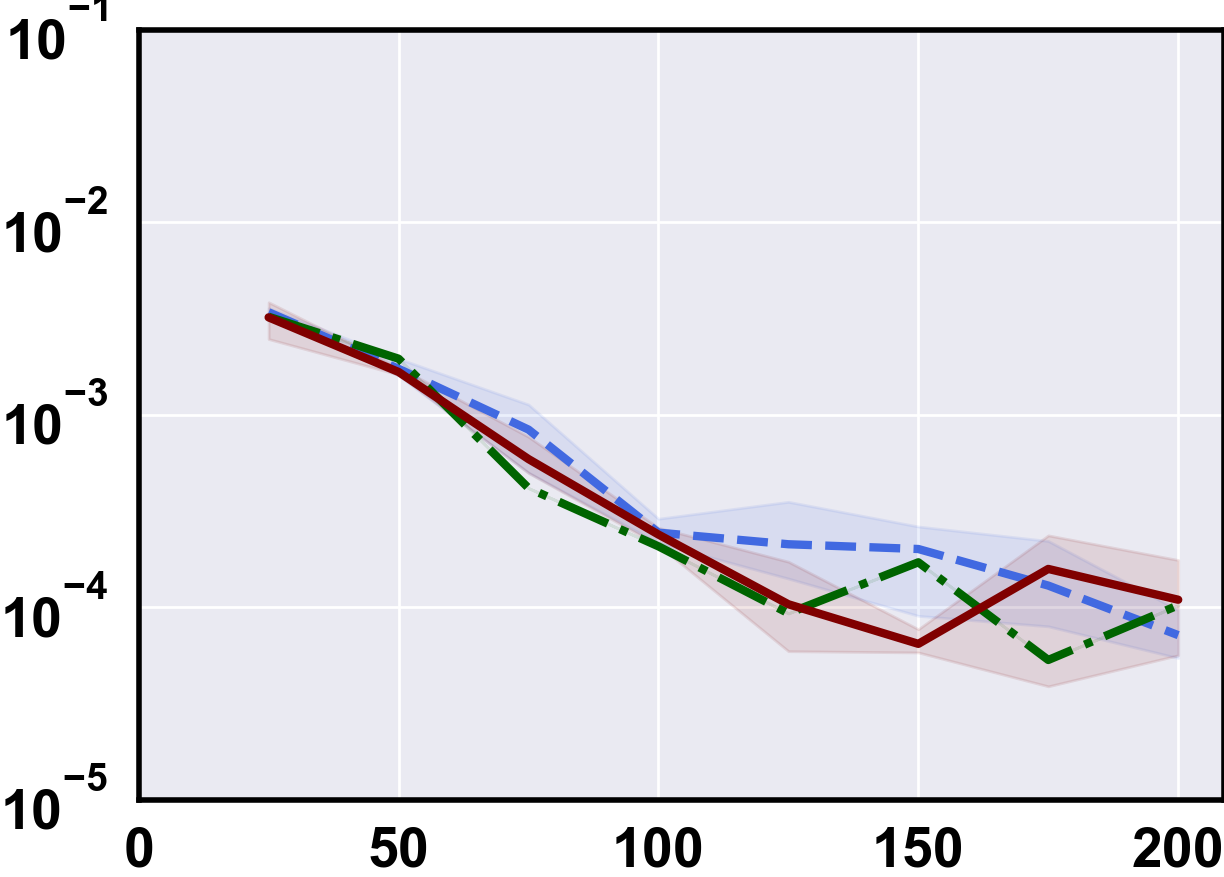}
    \end{subfigure}%
    \hfill%
    \begin{subfigure}{.32\linewidth}
        \centering
        \includegraphics[width=\linewidth]{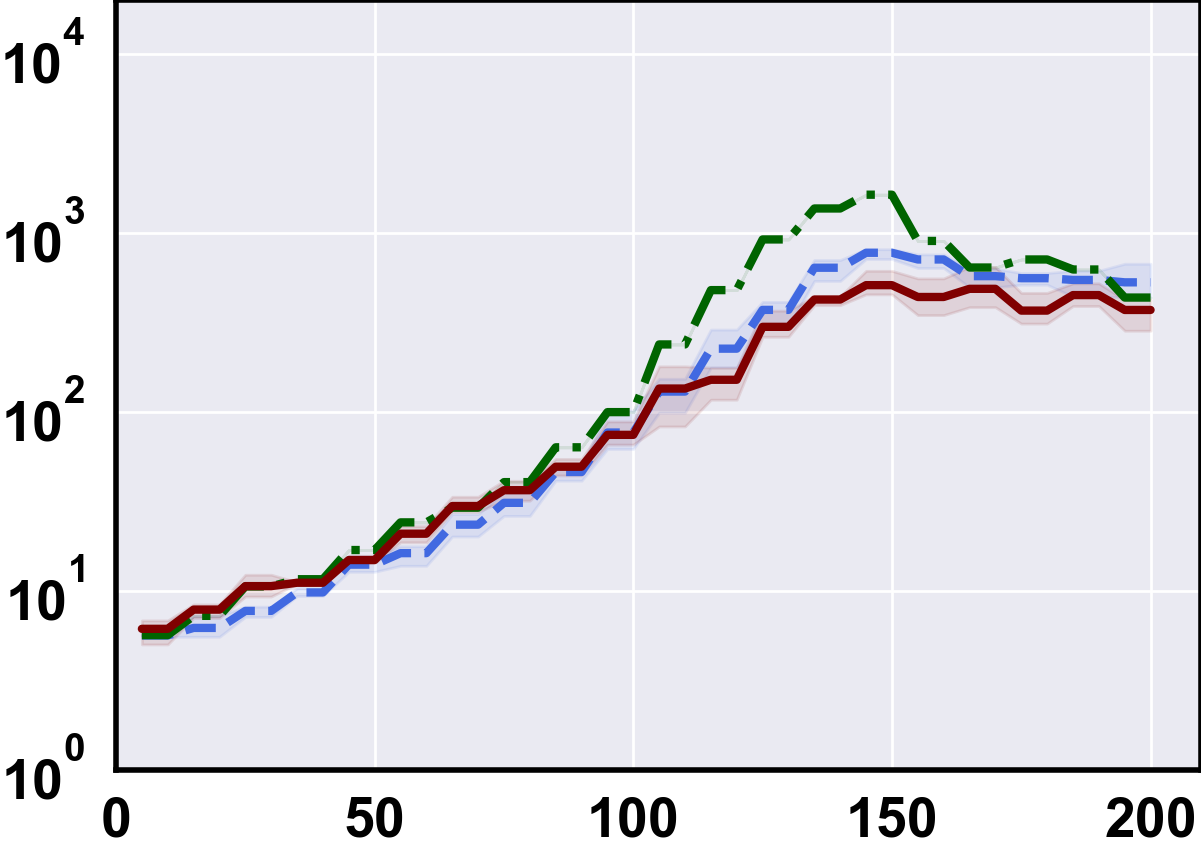}
    \end{subfigure}
    \end{minipage}
    
    \caption{Learning curves in the setting 2x3. The X-axis is in the thousands of training iterations. The shaded regions correspond to the min-max spread over three random seeds.}
    \label{fig:uni_2x3}
\end{figure}

\begin{figure}[t]
    \centering   
    \noindent\begin{minipage}{0.06\textwidth}
    \end{minipage}%
    \hfill%
    \begin{minipage}{0.94\textwidth}
    \centering
    \begin{subfigure}{.65\linewidth}
        \centering
        \includegraphics[width=\linewidth]{pics/legend.png}
    \end{subfigure}
    \end{minipage}
    
    \noindent\begin{minipage}{0.06\textwidth}
    $R_{max}$\\
    $10^{-3}$
    \end{minipage}%
    \hfill%
    \begin{minipage}{0.94\textwidth}
    \begin{subfigure}{.33\textwidth}
        \centering
        \caption{revenue}
        \includegraphics[width=\linewidth]{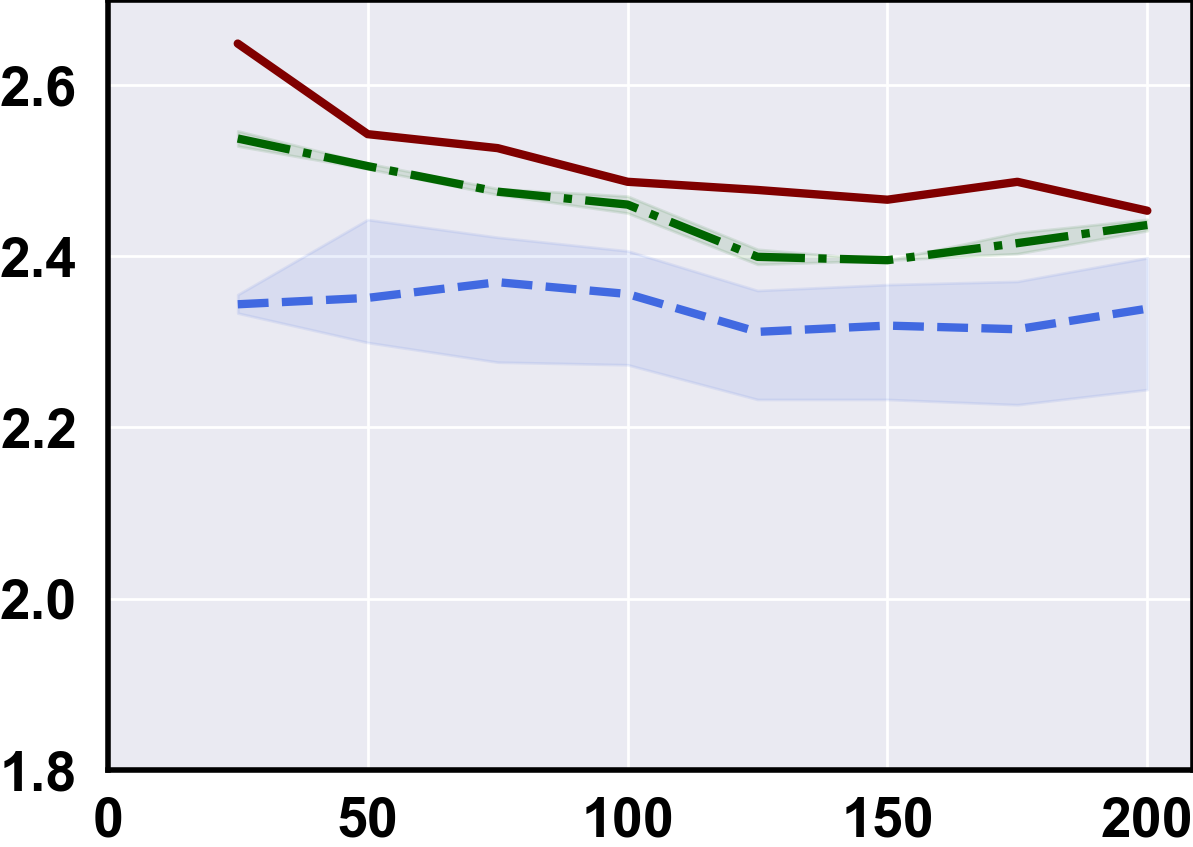}
    \end{subfigure}%
    \hfill%
    \begin{subfigure}{.33\textwidth}
        \centering
        \caption{regret}
        \includegraphics[width=\linewidth]{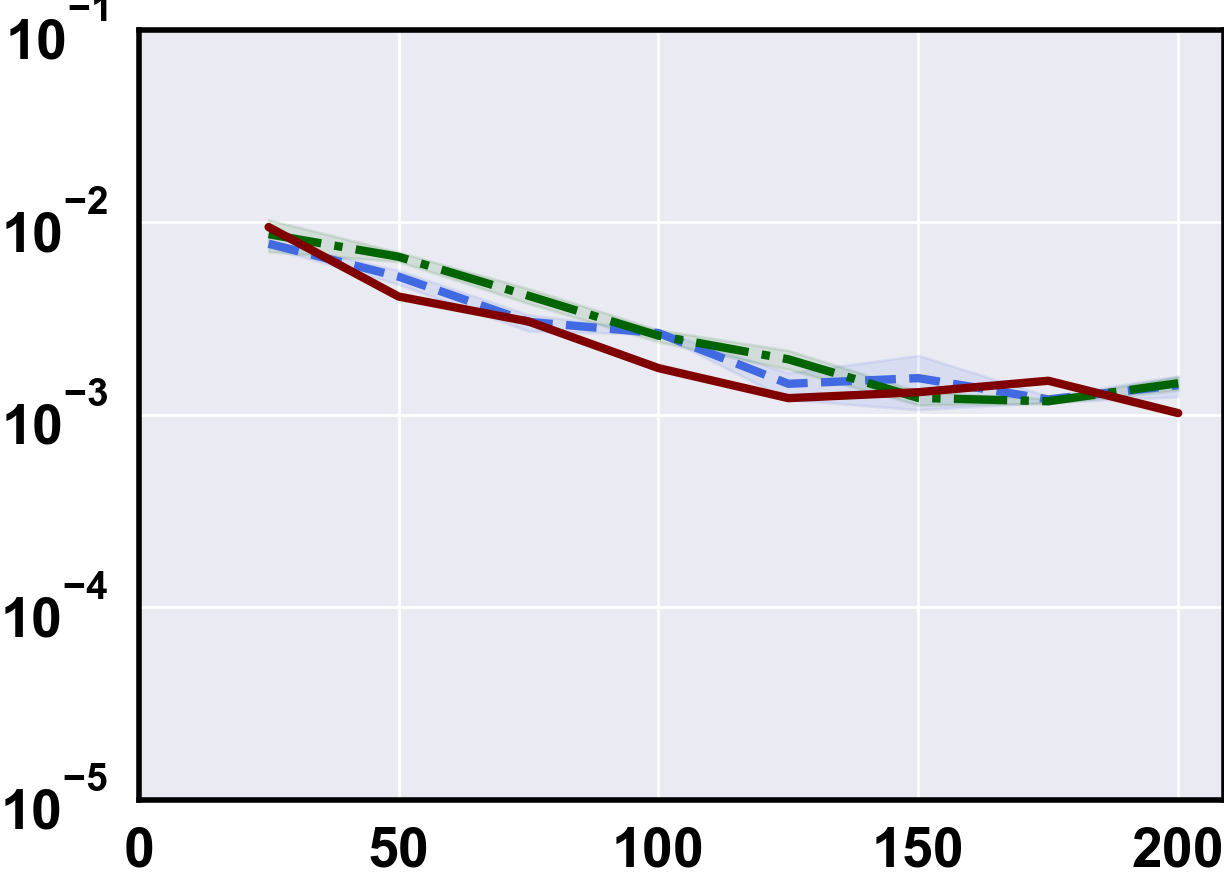}
    \end{subfigure}%
    \hfill%
    \begin{subfigure}{.32\textwidth}
        \centering
        \caption{penalty coefficient $\gamma$}
        \includegraphics[width=\linewidth]{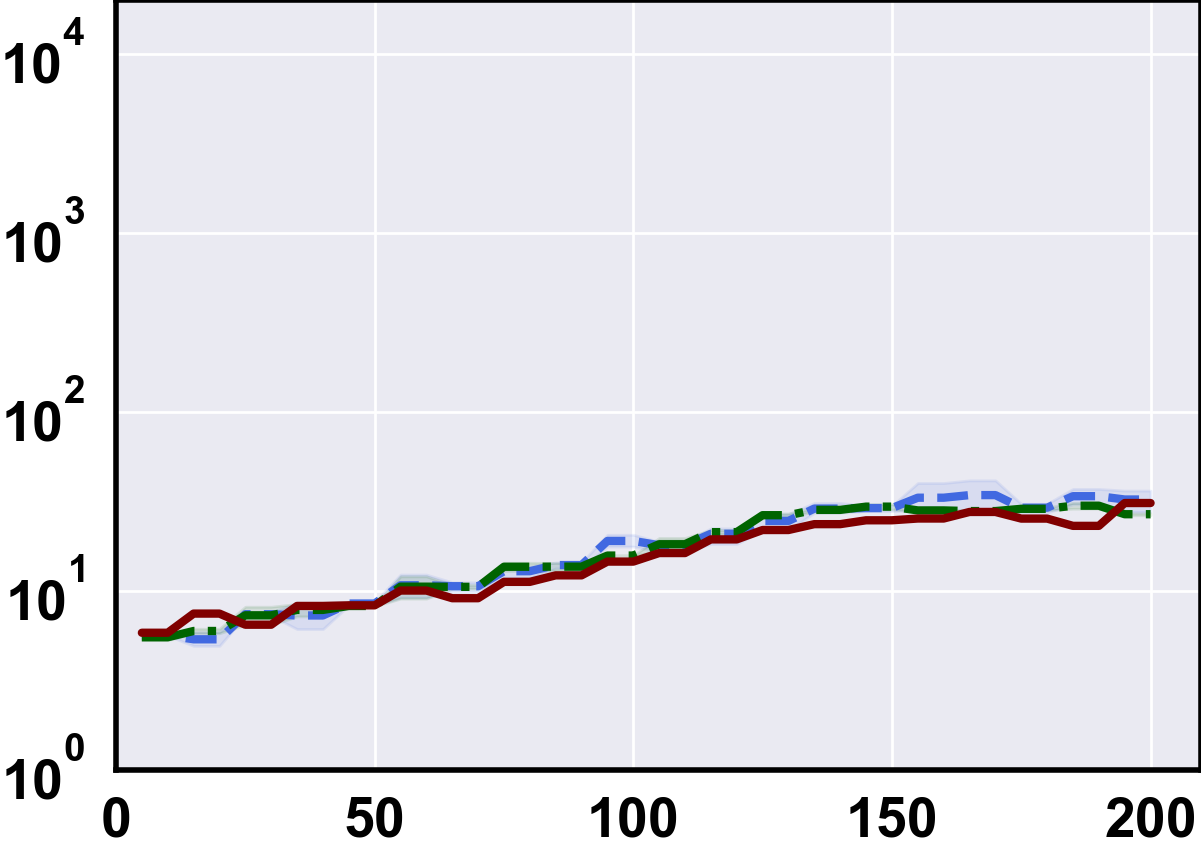}
    \end{subfigure}
    \end{minipage}

    \noindent\begin{minipage}{0.06\textwidth}
    $R_{max}$\\
    $10^{-4}$
    \end{minipage}%
    \hfill%
    \begin{minipage}{0.94\textwidth}
    \begin{subfigure}{.33\linewidth}
        \centering
        \includegraphics[width=\linewidth]{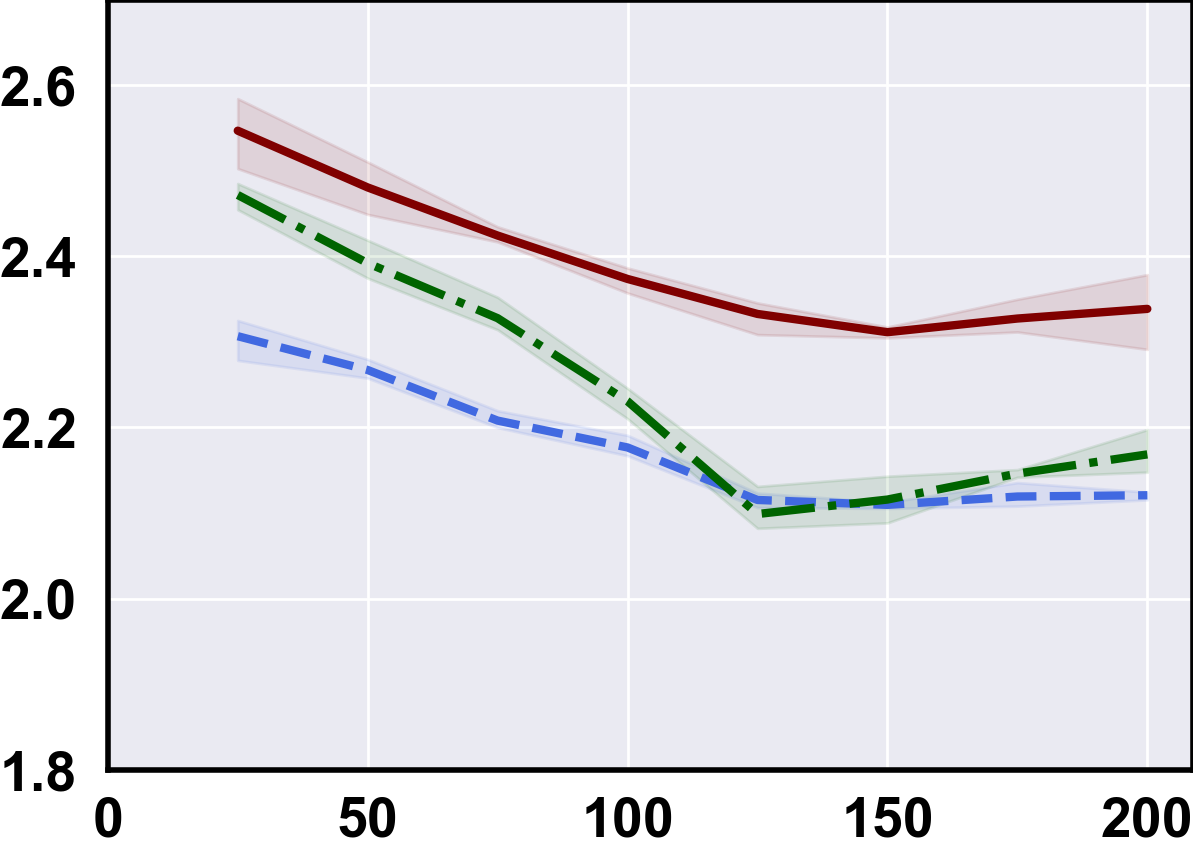}
    \end{subfigure}%
    \hfill%
    \begin{subfigure}{.33\linewidth}
        \centering
        \includegraphics[width=\linewidth]{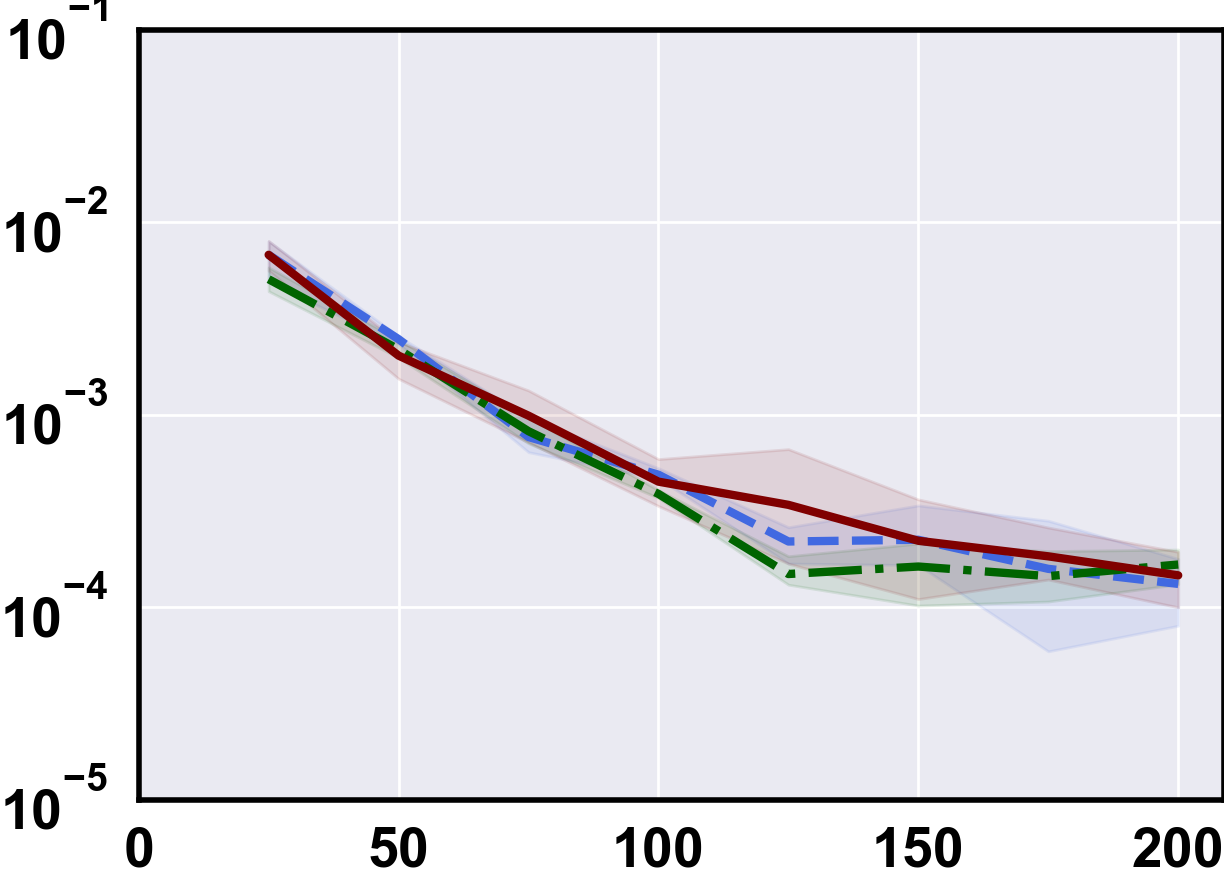}
    \end{subfigure}%
    \hfill%
    \begin{subfigure}{.32\linewidth}
        \centering
        \includegraphics[width=\linewidth]{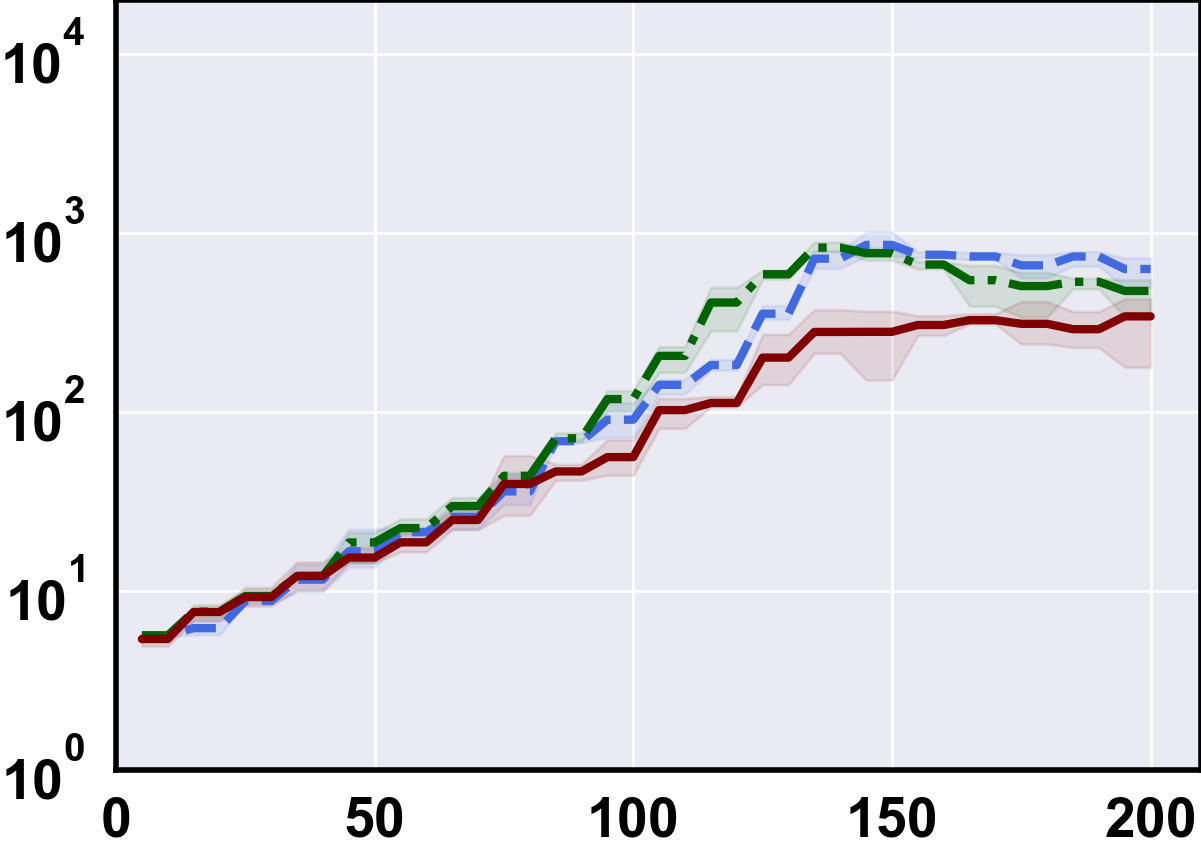}
    \end{subfigure}
    \end{minipage}

    \caption{Learning curves in the setting 2x5. The X-axis is in the thousands of training iterations. The shaded regions correspond to the min-max spread over three random seeds.}
    \label{fig:uni_2x5}
\end{figure}

\begin{figure}[t]
    \centering    
    \noindent\begin{minipage}{0.06\textwidth}
    \end{minipage}%
    \hfill%
    \begin{minipage}{0.94\textwidth}
    \centering
    \begin{subfigure}{.65\linewidth}
        \centering
        \includegraphics[width=\linewidth]{pics/legend.png}
    \end{subfigure}
    \end{minipage}
    
    \noindent\begin{minipage}{0.06\textwidth}
    $R_{max}$\\
    $10^{-3}$
    \end{minipage}%
    \hfill%
    \begin{minipage}{0.94\textwidth}
    \begin{subfigure}{.33\textwidth}
        \centering
        \caption{revenue}
        \includegraphics[width=\linewidth]{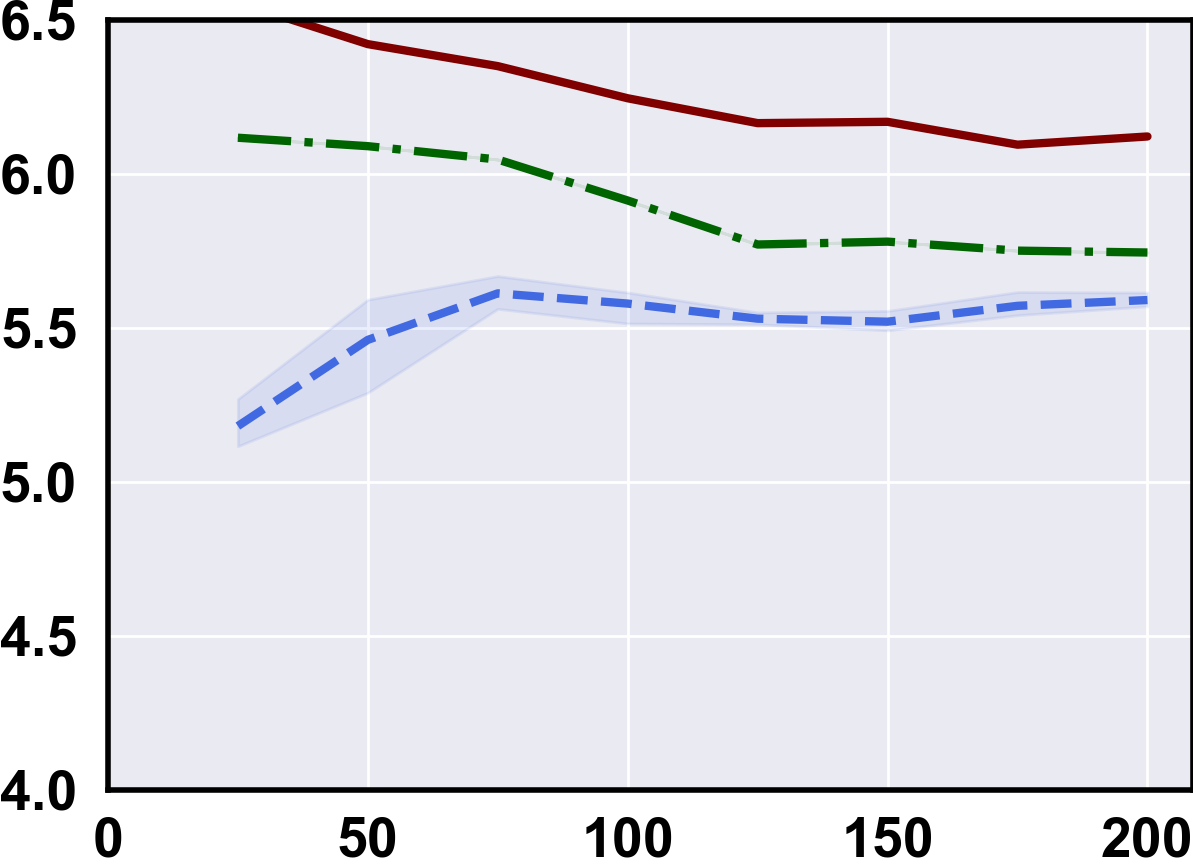}
    \end{subfigure}%
    \hfill%
    \begin{subfigure}{.33\textwidth}
        \centering
        \caption{regret}
        \includegraphics[width=\linewidth]{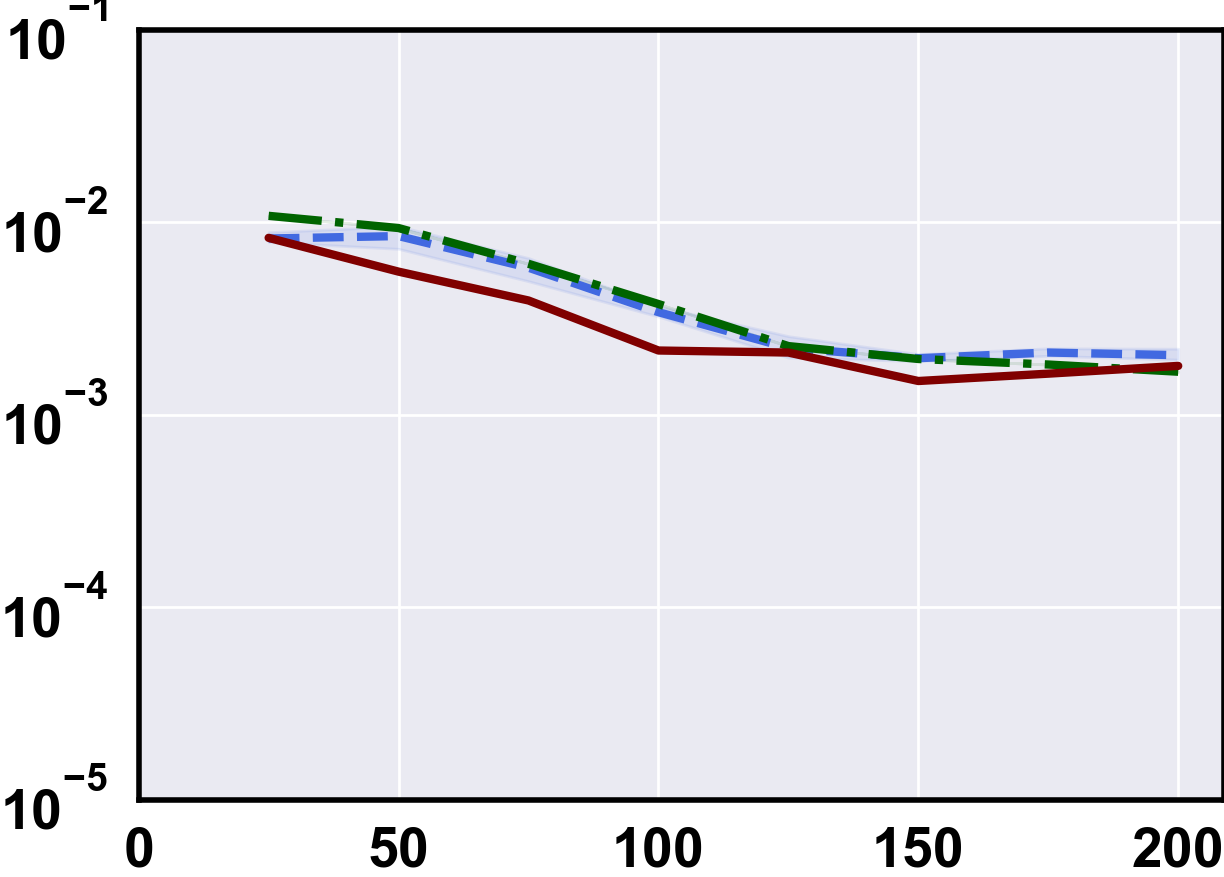}
    \end{subfigure}%
    \hfill%
    \begin{subfigure}{.32\textwidth}
        \centering
        \caption{penalty coefficient $\gamma$}
        \includegraphics[width=\linewidth]{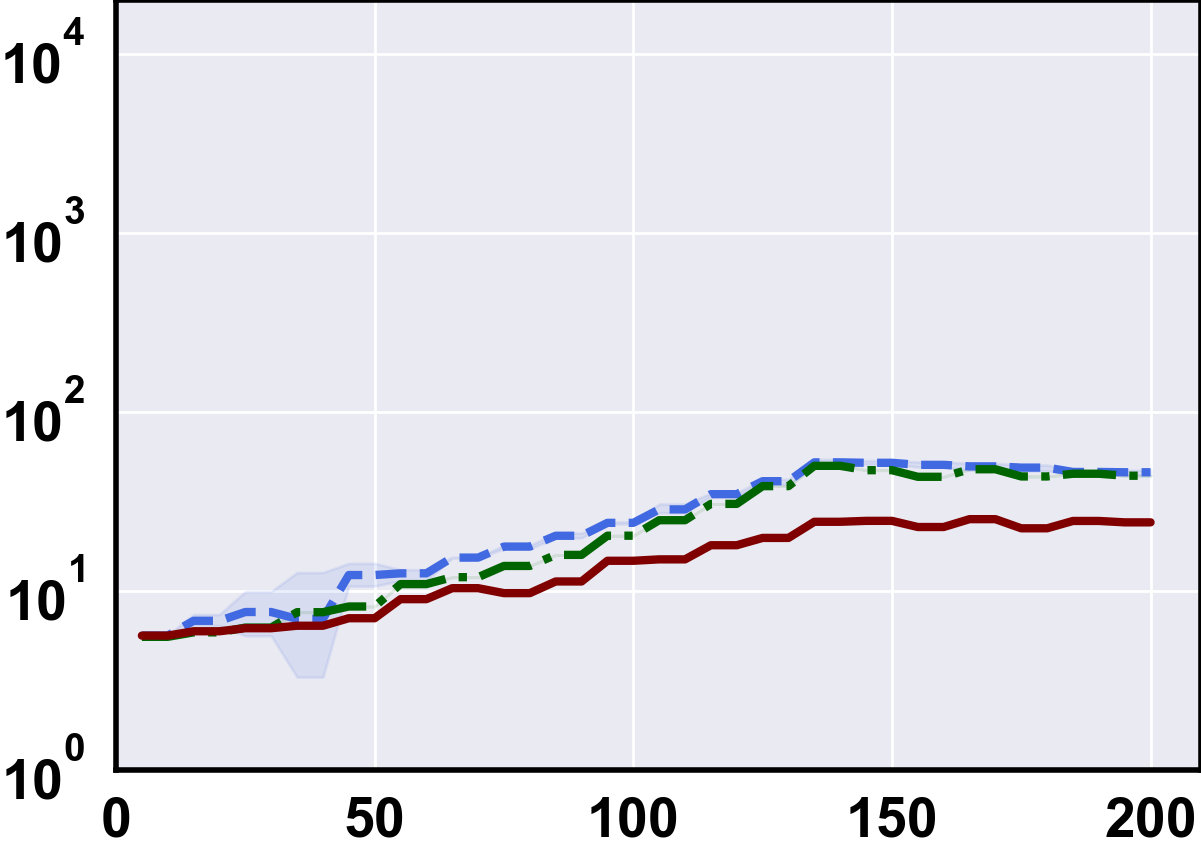}
    \end{subfigure}
    \end{minipage}

    \noindent\begin{minipage}{0.06\textwidth}
    $R_{max}$\\
    $10^{-4}$
    \end{minipage}%
    \hfill%
    \begin{minipage}{0.94\textwidth}
    \begin{subfigure}{.33\linewidth}
        \centering
        \includegraphics[width=\linewidth]{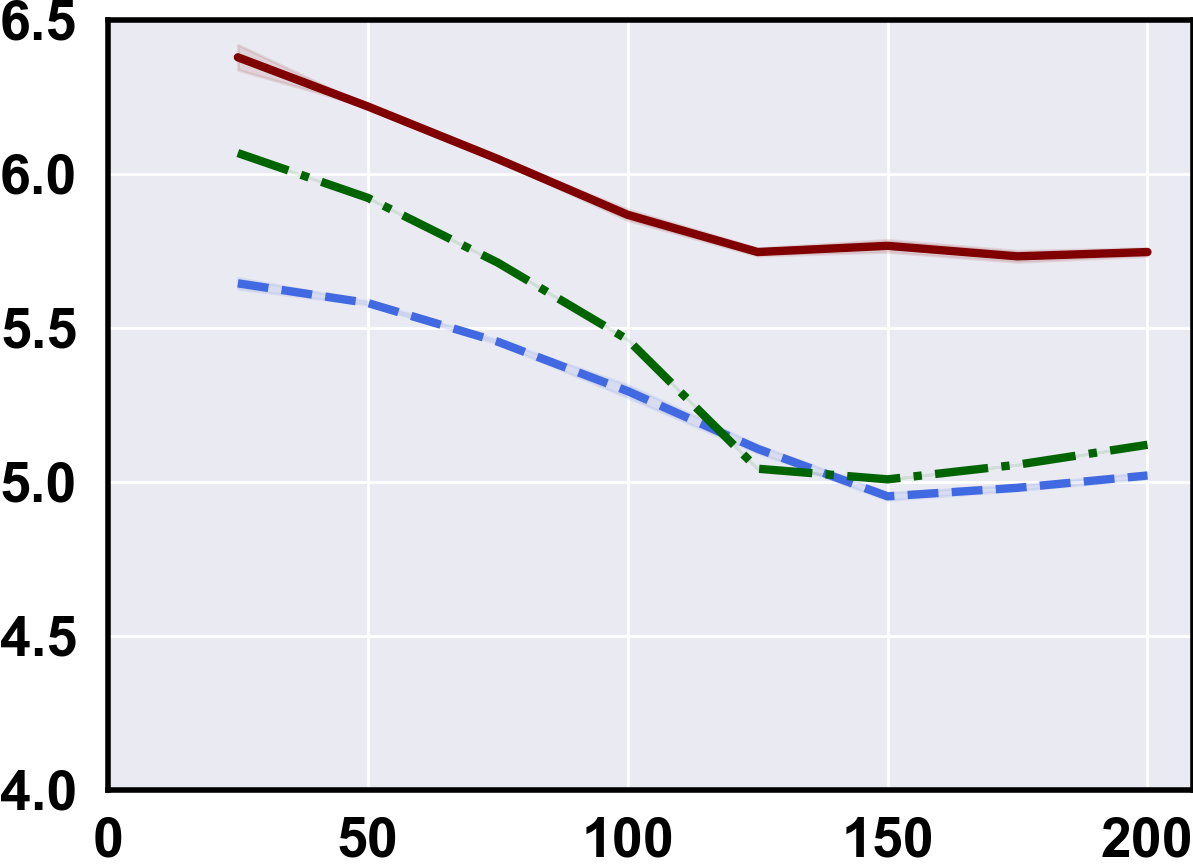}
    \end{subfigure}%
    \hfill%
    \begin{subfigure}{.33\linewidth}
        \centering
        \includegraphics[width=\linewidth]{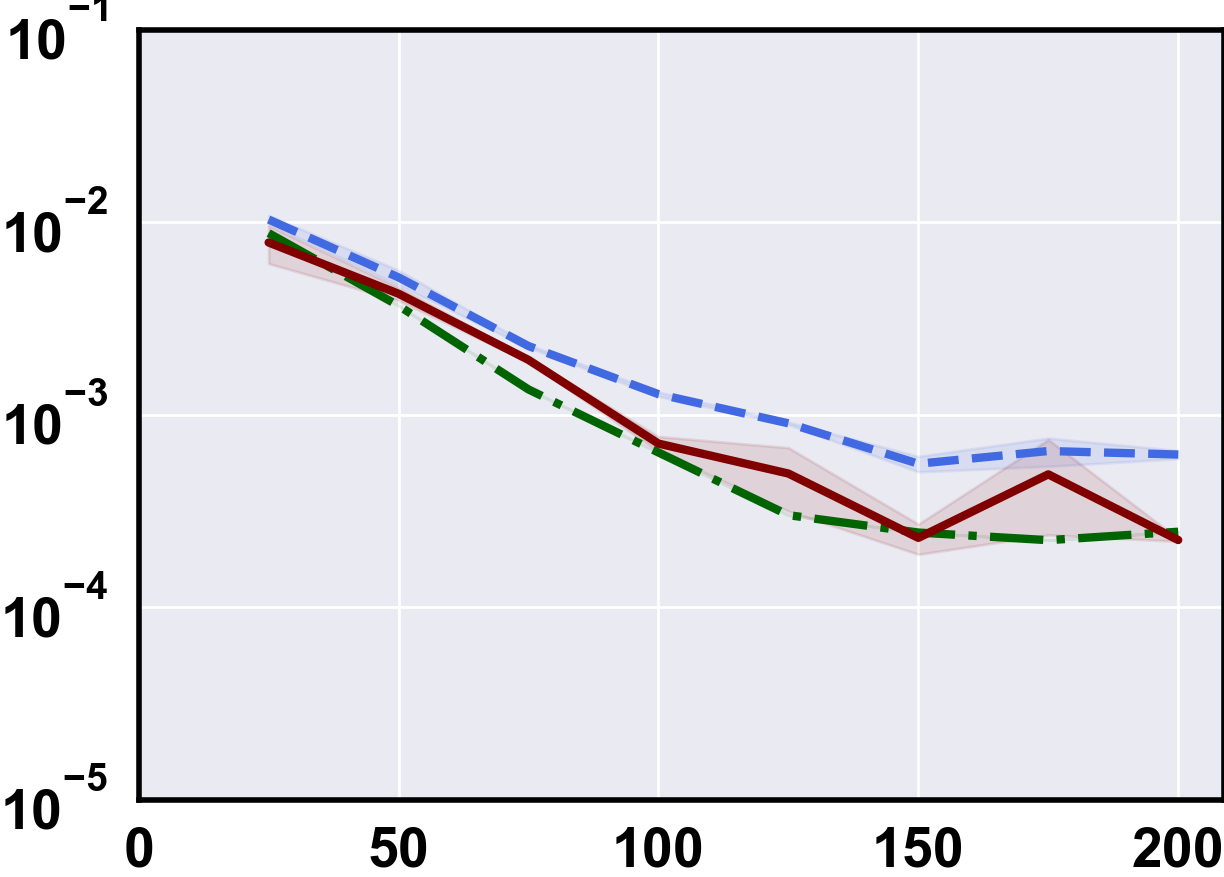}
    \end{subfigure}%
    \hfill%
    \begin{subfigure}{.32\linewidth}
        \centering
        \includegraphics[width=\linewidth]{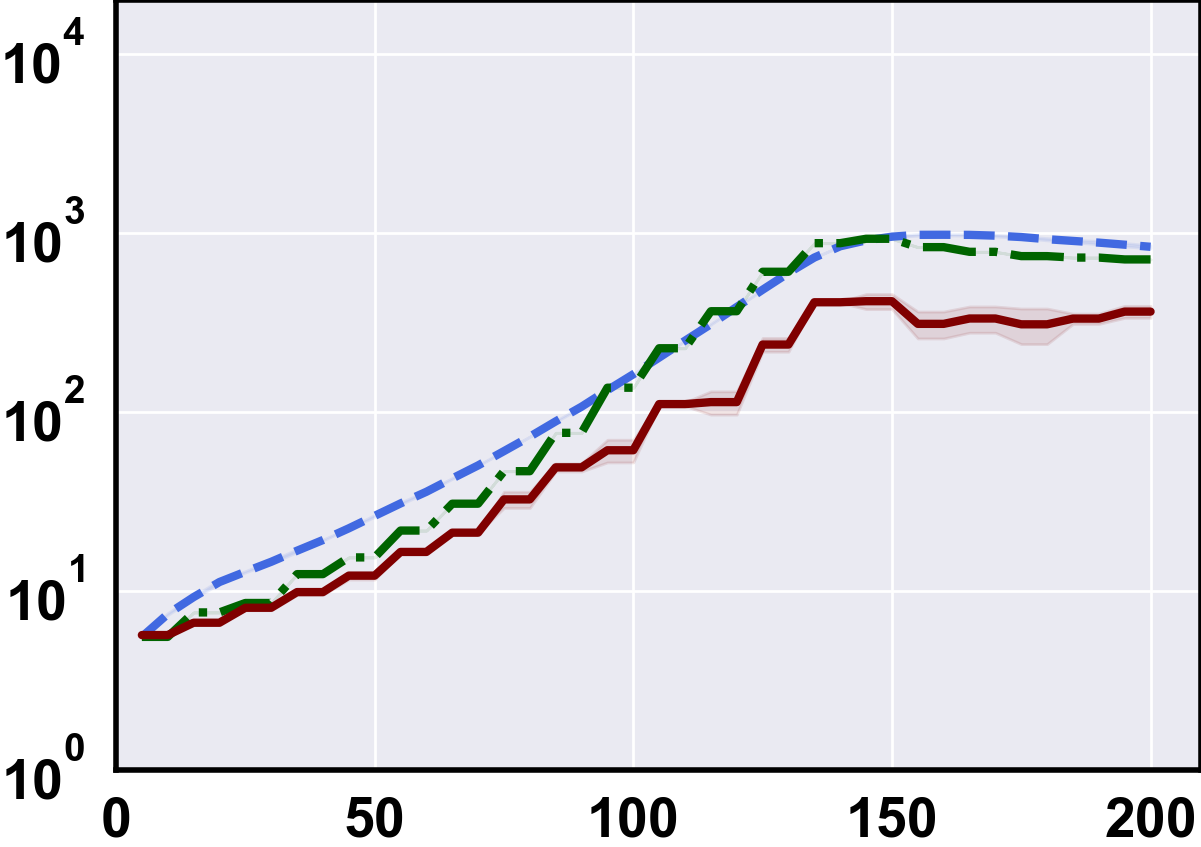}
    \end{subfigure}
    \end{minipage}
    
    \caption{Learning curves in the setting 3x10. The X-axis is in the thousands of training iterations. The shaded regions correspond to the min-max spread over three random seeds.}
    \label{fig:uni_3x10}
\end{figure}

We present the learning curves of the revenue, the regret, and the penalty coefficient $\gamma$ for settings \{1x2, 2x2, 2x3, 2x5, 3x10\} in Figures \ref{fig:uni_1x2}, \ref{fig:uni_2x2}, \ref{fig:uni_2x3}, \ref{fig:uni_2x5}, and \ref{fig:uni_3x10}, respectively. The results of the same experiments are reported in Table \ref{table:uni} in the main text.

Note that the shapes of learning curves are a consequence of our regret budget schedule. Specifically, we provide a higher regret budget at the beginning of the training so the network finds a solution with high revenue (that does not satisfy the desirable budget), and then we tighten the regret budget (which also causes the revenue to decrease). It can also be seen that both revenue and regret start flattening at the same time as the regret penalty coefficient $\gamma$ stops increasing. This happens approximately at 2/3 of the training time, in accordance with our choice of $R_{max}^{mult}$.

\subsection{Network distillation}\label{app:experiments_dist}

Here we elaborate on our validation procedure based on network distillation that was mentioned in Section \ref{sec:experiments_val} of the main text.

The distillation procedure is based on training the `student` network $w_s$ to approximate the predictions of a trained `teacher` network $w_t$. We apply this procedure in five settings \{1x2, 2x2, 2x3, 2x5, 3x10\} given $R_{max} = 10^{-3}$ to distill RegretFormer onto RegretNet, as well as to distill RegretFormer onto EquivariantNet. If the architecture of RegretFormer for some reason impairs its ability to approximate optimal misreports, a RegretNet or an EquivariantNet trained to closely mimic the predictions of a RegretFormer may find better misreports that produce higher regret values for the RegretFormer. Specifically, since the predictions of allocation and payment modules can respectively be treated as the categorical and the Bernoulli distributions, we train the student network to minimize the KL divergence from its predictions to the predictions of the teacher network. For example, to train the allocation module, the student minimizes $KL(g(w_t), g(w_s)) = \frac{1}{\left|B\right|} \sum_{l \in B} \sum_{i, j} z_{ij}(v^l; w_t) \cdot (\log(z_{ij}(v^l; w_t)) - \log(z_{ij}(v^l; w_s)))$, and likewise for the payment module. This approach was initially proposed in \citet{hinton2015distilling}. Furthermore, to satisfy DSIC on the whole support, the KL-divergence is minimized at both the true valuations $v^l$ and the approximate optimal misreports for each participant $v_i^{\prime l}(w_s)$, estimated by the student network as per usual via inner optimization (\ref{eq:loss_inner}).

We report the results in Table \ref{table:distillation}. Like in the cross-misreport validation procedure, we do not find any evidence that RegretFormer approximates regret worse than the alternative architectures. This can be seen in Table \ref{table:distillation} by comparing the teacher regret estimated on teacher misreports with the teacher regret estimated on student misreports: the latter is never substantially higher than the former.

As additional evidence towards the performance gap being genuine, in the distillation experiments both students achieve the same revenue as the teacher while consistently producing higher regret, up to a magnitude on the hardest 3x10 setting. This can have two explanations. First, the student networks get stuck in one of multiple local optimums, which prevents them from reaching lower regrets. Moreover, this consistently happens both when learning from scratch (since in Table \ref{table:uni} both RegretNet and EquivariantNet achieve lower revenue than RegretFormer for the same regret budgets) and when mimicking predictions of RegretFormer. Second, the better solutions with high revenue and low regret that can be found by RegretFormer are simply absent from the space of the mechanisms that can be represented by RegretNet and EquivariantNet.

\begin{table}[t]
\centering
\caption{Network distillation using networks from Table \ref{table:uni}, $R_{max}=10^{-3}$}
\label{table:distillation}
\begin{tabular}{lllcccc}
\toprule
setting & metric                      & misreports of & \multicolumn{2}{c}{RegretFormer $\rightarrow$ RegretNet}  & \multicolumn{2}{c}{RegretFormer $\rightarrow$ EquivariantNet} \\
        &                             &               & \multicolumn{1}{c}{teacher} & \multicolumn{1}{c}{student} & \multicolumn{1}{c}{teacher}   & \multicolumn{1}{c}{student}   \\ 
        \midrule
1x2     & \multicolumn{1}{c}{revenue} & - & 0.577 & 0.578 & 0.578 & 0.578 \\
 & regret & teacher & 0.00087 & 0.00098 & 0.00090 & 0.00120 \\
 & & student & 0.00051 & 0.00090 & 0.00024 & 0.00066 \\
        \midrule
2x2     & revenue                     & - & 0.912 & 0.912 & 0.913 & 0.917 \\
 & regret & teacher & 0.00057 & 0.00074 & 0.00057 & 0.00154 \\
 &  & student & 0.00068 & 0.00148 & 0.00054 & 0.00404 \\
        \midrule
2x3     & revenue                     & - & 1.412 & 1.412 & 1.414 & 1.419 \\
 & regret & teacher & 0.00097 & 0.00153 & 0.00093 & 0.00183 \\
 &  & student & 0.00129 & 0.00415 & 0.00090 & 0.00400 \\
        \midrule
2x5     & revenue                     & - & 2.439 & 2.436 & 2.440 & 2.449 \\
 & regret & teacher & 0.00112 & 0.00125 & 0.00110 & 0.00178 \\
 &  & student & 0.00106 & 0.00416 & 0.00064 & 0.00577 \\
        \midrule
3x10    & revenue                     & - & 6.153 & 6.169 & 6.155 & 6.163 \\
 & regret & teacher & 0.00238 & 0.00386 & 0.00237 & 0.00298 \\
 &  & student & 0.00258 & 0.02713 & 0.00220 & 0.01626 \\
        \bottomrule
\end{tabular}
\end{table}

\subsection{Out-of-setting generalization}\label{app:experiments_out_uni}

In these experiments, we investigate how well the architectures generalize to unseen settings. It's clear that RegretNet cannot be applied out-of-domain since its layers rely on constant input size, so we compare our network with EquivariantNet. The networks are trained in five settings \{1x2, 2x2, 2x3, 2x5, 3x10\} and then tested in all settings but the setting used for training. 

The results are reported in Table \ref{table:uni_out}. Both networks look promising when the number of objects varies and the number of bidders remains constant. However, generalization to the settings where the number of bidders varies is poor for both networks due to complex interactions between the participants. Similar results were observed by \citet{rahme2021permutation}. Still, when the validation setting has the number of participants same as or less than the training setting, RegretFormer usually outperforms EquivariantNet by either achieving higher revenue or lower regret.

\begin{table}[t]
\centering
\caption{Out-of-setting generalization}
\label{table:uni_out}
\begin{tabular}{llllll}
\toprule
Training & Validation & \multicolumn{2}{c}{EquivariantNet} & \multicolumn{2}{c}{RegretFormer} \\
setting              &  setting              & revenue          & regret          & revenue    & regret             \\
\midrule
1x2              & 2x2                & 0.690            & 0.04403         & 0.669      & 0.03776   \\
                 & 2x3                & 1.084            & 0.07074         & 1.056      & 0.07167            \\
                 & 2x5                & 1.917            & 0.12465         & 1.863      & 0.13151            \\
                 & 3x10               & 4.308            & 0.29291         & 4.197      & 0.29106            \\
\midrule
2x2              & 1x2                & 0.695            & 0.13343         & 0.768      & 0.21671            \\
                 & 2x3                & 1.350            & 0.00071         & 1.412      & 0.01221            \\
                 & 2x5                & 2.307            & 0.03169         & 2.402      & 0.03250            \\
                 & 3x10               & 5.156            & 0.55869         & 4.943      & 0.32002            \\
\midrule
2x3              & 1x2                & 0.686            & 0.14900         & 0.775      & 0.20051            \\
                 & 2x2                & 0.875            & 0.00116         & 0.904      & 0.00137            \\
                 & 2x5                & 2.318            & 0.00615         & 2.432      & 0.00938            \\
                 & 3x10               & 5.271            & 0.37967         & 4.929      & 0.27183            \\
\midrule
2x5              & 1x2                & 0.743            & 0.19830         & 0.816      & 0.23945            \\
                 & 2x2                & 0.900            & 0.00066         & 0.903      & 0.00072            \\
                 & 2x3                & 1.401            & 0.00103         & 1.415      & 0.00103            \\
                 & 3x10               & 5.517            & 0.24757         & 4.884      & 0.28009            \\
\midrule
3x10             & 1x2                & 0.552            & 0.53700         & 0.801      & 0.18780            \\
                 & 2x2                & 0.693            & 0.19767         & 1.053      & 0.05893            \\
                 & 2x3                & 1.099            & 0.33665         & 1.611      & 0.07595            \\
                 & 2x5                & 1.936            & 0.59959         & 2.754      & 0.10181           \\
\bottomrule
\end{tabular}
\end{table}

\begin{table}[t]
\centering
\caption{Out-of-multi-setting generalization}
\label{table:multi_out}
\begin{tabular}{lllll}
\toprule
\multirow{2}{*}{Setting} & \multicolumn{2}{c}{RegretNet} & \multicolumn{2}{c}{RegretFormer} \\
                         & revenue        & regret       & revenue         & regret        \\
\midrule
average                     & 2.733          & 0.0662        & 2.734           & 0.00478         \\
\midrule
2x4                      & 2.115          & 0.083        & 1.943           & 0.00368         \\
2x6                      & 3.166          & 0.078        & 3.012           & 0.00592         \\
3x3                      & 1.743          & 0.010        & 1.718           & 0.00252         \\
3x5                      & 2.918          & 0.016        & 2.910           & 0.00313         \\
3x7                      & 3.722          & 0.144        & 4.086           & 0.00866       \\ 
\bottomrule
\end{tabular}
\end{table}

\begin{table}[t]
\centering
\caption{Cross-misreport regret estimates in the multi-setting. The highest regret for a network is highlighted with bold. Notice that EquivariantNet poorly estimates misreports as its regret on misreports of RegretFormer consistently higher than on its own misreports.}
\label{table:cross_misreport_multi_big}
\begin{tabular}{lllll}
\toprule
setting & regret of & \multicolumn{3}{c}{misreports of} \\
        &               & RegretNet    & EquivariantNet & RegretFormer \\
\midrule
\multirow{3}{*}{2x3} & RegretNet      & \textbf{0.00305} & 0.00083        & 0.00134          \\
                     & EquivariantNet & 0.00514          & 0.00258        & \textbf{0.00554} \\
                     & RegretFormer   & 0.00128          & 0.00105        & \textbf{0.00246} \\
\midrule
\multirow{3}{*}{2x4} & RegretNet      & \textbf{0.00341} & 0.00060        & 0.00128          \\
                     & EquivariantNet & 0.00609          & 0.00273        & \textbf{0.00925} \\
                     & RegretFormer   & 0.00154          & 0.00106        & \textbf{0.00317} \\
\midrule
\multirow{3}{*}{2x5} & RegretNet      & \textbf{0.00362} & 0.00050        & 0.00115          \\
                     & EquivariantNet & 0.00726          & 0.00309        & \textbf{0.01270} \\
                     & RegretFormer   & 0.00188          & 0.00113        & \textbf{0.00391} \\
\midrule
\multirow{3}{*}{2x6} & RegretNet      & \textbf{0.00425} & 0.00065        & 0.00131          \\
                     & EquivariantNet & 0.00873          & 0.00337        & \textbf{0.01597} \\
                     & RegretFormer   & 0.00248          & 0.00118        & \textbf{0.00439} \\
\midrule
\multirow{3}{*}{2x7} & RegretNet      & \textbf{0.00457} & 0.00048        & 0.00111          \\
                     & EquivariantNet & 0.00961          & 0.00356        & \textbf{0.01951} \\
                     & RegretFormer   & 0.00242          & 0.00122        & \textbf{0.00481} \\
\midrule
\multirow{3}{*}{3x3} & RegretNet      & \textbf{0.00322} & 0.00070        & 0.00101          \\
                     & EquivariantNet & 0.00254          & 0.00189        & \textbf{0.00358} \\
                     & RegretFormer   & 0.00110          & 0.00083        & \textbf{0.00251} \\
\midrule
\multirow{3}{*}{3x4} & RegretNet      & \textbf{0.00264} & 0.00042        & 0.00075          \\
                     & EquivariantNet & 0.00354          & 0.00214        & \textbf{0.00508} \\
                     & RegretFormer   & 0.00138          & 0.00087        & \textbf{0.00336} \\
\midrule
\multirow{3}{*}{3x5} & RegretNet      & \textbf{0.00277} & 0.00032        & 0.00058          \\
                     & EquivariantNet & 0.00426          & 0.00265        & \textbf{0.00709} \\
                     & RegretFormer   & 0.00167          & 0.00096        & \textbf{0.00421} \\
\midrule
\multirow{3}{*}{3x6} & RegretNet      & \textbf{0.00340} & 0.00030        & 0.00058          \\
                     & EquivariantNet & 0.00501          & 0.00277        & \textbf{0.00916} \\
                     & RegretFormer   & 0.00171          & 0.00181        & \textbf{0.00476} \\
\midrule
\multirow{3}{*}{3x7} & RegretNet      & \textbf{0.00430} & 0.00027        & 0.00054          \\
                     & EquivariantNet & 0.00610          & 0.00326        & \textbf{0.01101} \\
                     & RegretFormer   & 0.00198          & 0.00101        & \textbf{0.00553} \\
\bottomrule
\end{tabular}
\end{table}

\subsection{Cross-misreport validation in multi-settings}\label{app:experiments_val_cross_multi}

In section \ref{sec:experiments_multi} in the main text, we have mentioned performing cross-misreport validation in multi-settings. Table \ref{table:cross_misreport_multi_big} presents full results of this experiment. It is evident that both RegretNet and RegretFormer approximate the optimal misreports adequately since each estimates the highest regret on own misreports. In contrast, EquivariantNet poorly approximates misreports and underestimates regret in multi-settings since both RegretNet and RegretFormer find better misreports for this network.

\subsection{Out-of-multi-setting generalization}\label{app:experiments_out_multi}

We define two subsets of settings to respectively train and validate networks:

\begin{enumerate}
    \item \emph{Train}: \(S_{train}\)=\{2x3, 2x5, 2x7, 3x4, 3x6\}.
    \item \emph{Test}: \(S_{test}\)=\{2x4, 2x6, 3x3, 3x5, 3x7\}.
\end{enumerate}

We compare how networks generalize to unseen settings when trained in the multi-setting regime. To this end, we train RegretNet and RegretFormer on $S_{train}$ and then validate them on $S_{test}$. Because EquivariantNet poorly underestimates regret in multi-settings (see Section \ref{sec:experiments_multi} and \ref{app:experiments_val_cross_multi}), we do not include it for comparison.

The resulting revenue and regret values are reported in Table \ref{table:multi_out}. In all out-of-domain settings RegretNet produced poor results. On average, its regret is more than an order of magnitude larger compared to RegretFormer. In contrast, our approach stably generalizes to all unseen settings while keeping regret low. Remarkably, its revenue is as high and its regret is as low as in the experiments where all constant-sized settings are available during training (Table \ref{table:multi_big}). The superiority of RegretFormer over RegretNet is especially prominent in the 3x7 setting where our network achieves a larger revenue while producing 16 times as little regret.

\subsection{Training time}\label{app:experiments_time}

\begin{table}[t]
\centering
\caption{Average wall-clock training time, hours}
\begin{tabular}{lllll}
\toprule
setting & RegretNet & EquivariantNet & RegretFormer \\
\midrule
1x2 & 2.5 & 5.2 & 12.5 \\
2x2 & 2.5 & 5.3 & 12.0 \\
2x3 & 2.5 & 8.4 & 11.0 \\
2x5 & 3.1 & 9.8 & 28.7 \\
3x10 & 4.7 & 11.2 & 82.1 \\
\bottomrule
\end{tabular}
\label{tab:time}
\end{table}

We report the average wall-clock training time in hours in Table \ref{tab:time}. It takes longer to train RegretFormer than the baselines for two reasons. First, to perform optimally RegretFormer requires a bigger network with more parameters than baselines, especially in the bigger settings. Please see Table \ref{table:size} for summarized sizes of the three neural architectures in all settings. Second, the attention layers have quadratic $O(n^2)$ complexity (where $n$ is the number of items or participants). If training time is an issue, the attention layers can be replaced with one of multiple modifications that have $O(n \log n)$ or $O(n)$ complexity \cite{shen2021efficient,kitaev2020reformer,wang2020linformer,katharopoulos2020transformers}.

\begin{table}[t]
\centering
\caption{Number of parameters. In preliminary experiments, we found that RegretNet and EquivariantNet do not benefit from increasing the network sizes past what was used in the respective papers, whereas RegretFormer requires more parameters to perform optimally.}
\label{table:size}
\begin{tabular}{llll}
\toprule
setting & RegretNet & EquivariantNet & RegretFormer \\
\midrule
1x2     & 21305     & 4546           & 12705        \\
2x2     & 22008     & 4546           & 49985        \\
2x3     & 22711     & 12802          & 49985        \\
2x5     & 84717     & 16930          & 362753       \\
3x10    & 91343     & 16930          & 362753       \\
\bottomrule 
\end{tabular}
\end{table}

\end{document}